\documentclass[runningheads]{llncs}
\usepackage{graphicx}
\usepackage{comment}
\usepackage{amsmath,amssymb} % define this before the line numbering.
\usepackage{color}
\usepackage{arydshln}
\usepackage{adjustbox}
\usepackage{multirow}
\usepackage{flafter}
\usepackage{subcaption}
\usepackage[innercaption]{sidecap}
\sidecaptionvpos{figure}{t}
\usepackage{verbatimbox}
\usepackage[width=122mm,left=12mm,paperwidth=146mm,height=193mm,top=12mm,paperheight=217mm]{geometry}
\newcommand{\Fig}{Fig. }
\newcommand{\Figure}{Figure }
\newcommand{\Tab}{Table }
\newcommand{\Supp}[1]{Supplement #1}
\newcommand{\Sec}{Section }

\newcommand{\aar}{}
\newcommand{\prg}[1]{\par\noindent{{\textbf{#1}}}}

\newcommand{\argmax}{\mathop{\rm arg~max}\limits}

\usepackage[skip=1pt,font=small,labelfont=bf,labelsep=period]{caption}

\begin{document}
% \linenumbers
\pagestyle{headings}
\mainmatter
\def\ECCVSubNumber{2487}  % Insert your submission number here

%\begin{comment}
\titlerunning{Irregularly Tabulated MLP} 
\authorrunning{Y.Sekikawa et al.} 
%\end{comment}
%******************

%%%%%%%%% TITLE
\title{Irregularly Tabulated MLP\\ for Fast Point Feature Embedding}
\author{Yusuke Sekikawa and Teppei Suzuki}
% \\
% {\tt\small \{ysekikawa,tsuzuki\}@d-itlab.co.jp}
% }
\institute{DENSO IT Laboratory}

\maketitle
%%%%%%%%% ABSTRACT
\begin{abstract}
Aiming at drastic speedup for point-feature embeddings at test time, we propose a new framework that uses a pair of multi-layer perceptrons (MLP) and a lookup table (LUT) to transform point-coordinate inputs into high-dimensional features.
When compared with PointNet’s \cite{qi2017pointnet} feature embedding part realized by MLP that requires millions of dot products, the proposed framework at test time requires no such layers of matrix-vector products but requires only looking up the nearest entities from the tabulated MLP followed by interpolation, defined over discrete inputs on a 3D lattice that is substantially arranged irregularly.
We call this framework ``LUTI-MLP: LUT Interpolation MLP" that provides a way to train end-to-end irregularly tabulated MLP coupled to a LUT in a specific manner without the need for any approximation at test time.
LUTI-MLP also provides significant speedup for Jacobian computation of the embedding function wrt global pose coordinate on Lie algebra $\mathfrak{se}(3)$ at test time, which could be used for point-set registration problems.
After extensive evaluation using the ModelNet40 \cite{wu20153d}, we confirmed that the LUTI-MLP even with a small (e.g., $4^3$) lattice yields performance comparable to that of the MLP while achieving significant speedup: $100\times$ for the embedding, $12\times$ for the approximate Jacobian, and $860\times$ for the canonical Jacobian. 
\end{abstract}
% \begin{abstract}
% No review required.
% \end{abstract}
% \vspace{7.5cm}

\begin{figure*}[ht!]
\begin{center}
\includegraphics[width=1\columnwidth]{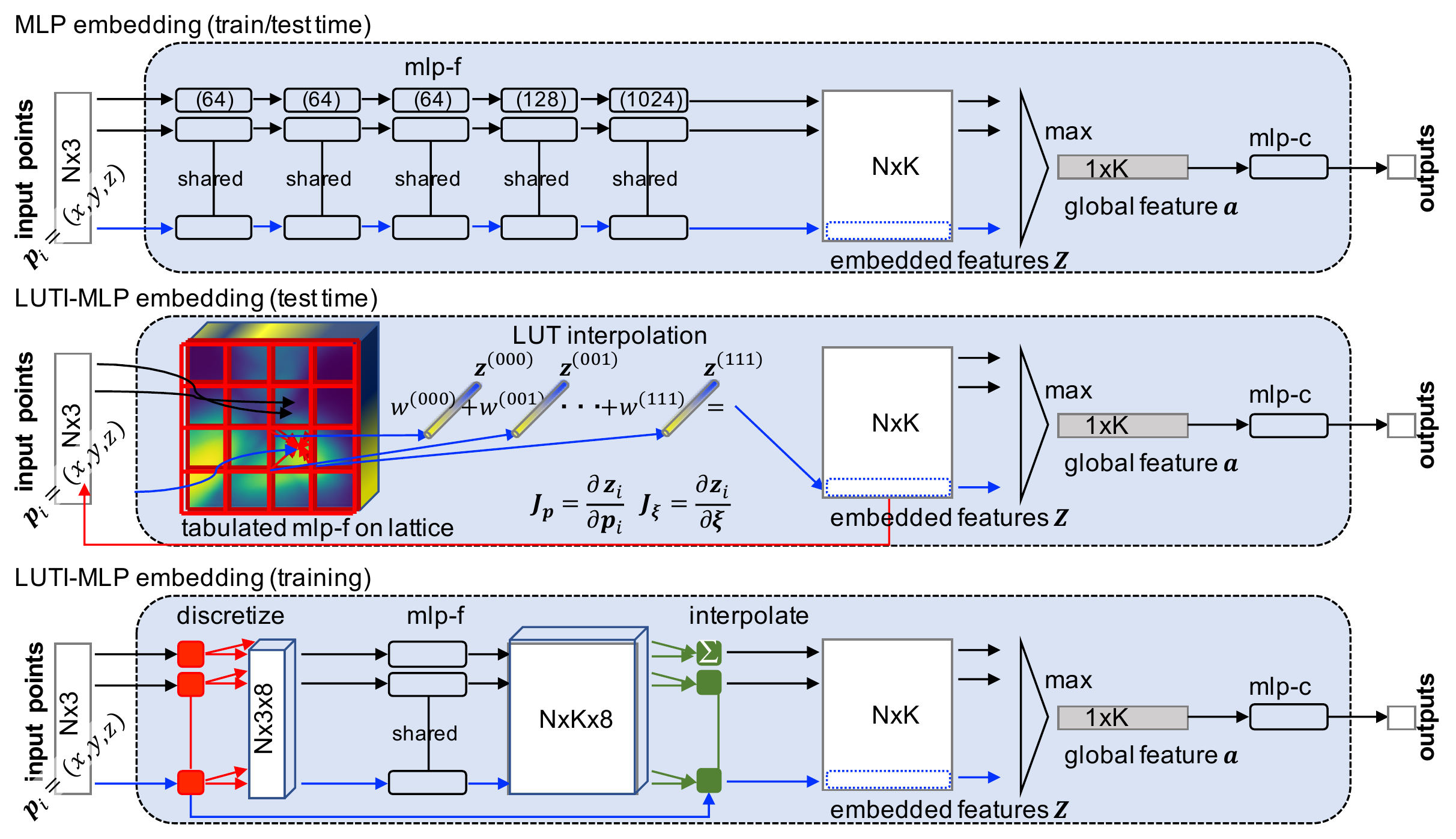}
\end{center}
\caption{
\label{fig:overview}
\textbf{LUTI-MLP embedding compared with MLP embedding for object
classification.}
Top: MLP embedding used in PointNet \cite{qi2017pointnet}.
Middle/Bottom: LUTI-MLP embedding at testing and training.
The network structure of LUTI-MLP is asymmetric at training and testing. 
At testing, 8 basis functions $\mathbf{\bar{z}}_i^{(j)}$ pre-computed from 8 discrete neighborhood points on the lattice $\mathbb{D}^3$ are linearly interpolated (uniform or irregular trilinear interpolation) to compute the embedding vector for input $\mathbf{p}_i=[x,y,z]$ in $\mathbb{R}^3$, 
as  $\mathbf{z}_{i}=\sum w_i^{(j)}\mathbf{\bar{z}}_i^{(j)}$.
LUTI-MLP significantly speeds up the embedding by eliminating the vector-matrix multiplication of MLP.
Furthermore, the Jacobian of the embedding vector wrt input point coordinate can also be computed efficiently (red line) for test time optimization, e.g., point set registration.
At training, the basis function $\mathbf{\bar{z}}_i^{(j)}= \phi_D(\mathbf{\bar{p}})$ is trained as an MLP.
% in an end-to-end manner in the same way as at the testing
% T-Net \cite{qi2017pointnet}  is omitted for clarity.
}
\end{figure*}
% such as from 3D point cloud,
%  look-up table interpolation 
% LUTI is comprised from differentiable discretization interpolation layers and it can be applied for any PointNet like embedding MLP. 
% The discrete set of the embedding feature is pre-computed for fast computation.
% LUTI module Look-up talble vertion of PoinNet is realized by
% our  discretization/interpolation module. 
% The entire network with LUT module is trainable in end-to-end manner.% The output using LUT is exactly the same as the one using MLP, since the the same LUTI is used for training and inference.
% By plugging LUTI before and after the point-wise embedding multi-layer perceptron (MLP) in training, the embedding operation is accelerated for more than 50 time in inference using LUT.

\section{Introduction}
Three-dimensional (3D) deep learning, which processes a set of points such as captured by depth sensors, has large practical applications: factory automation, vehicle automation, surveillance, and AR/VR.
These applications often require real-time processing on an edge device that has limited computational power and battery capacity.
Conventionally, these data points are processed using convolutional architecture \cite{tran2015learning,maturana2015voxnet,roynard2018classification,cciccek20163d,choy20163d,riegler2017octnet} after converting the sparse inputs into dense representation.
To reduce the information loss during the voxelization, fine grained input must be processed; however, the computational and memory requirements prohibit scaling to higher resolution.

Recently, approaches aiming at directly processing sparse points without voxelization have received increasing attention because of the efficiency they offer in terms of memory footprint and computation.
These approaches are applied for a range of tasks, such as object detection \cite{qi2019deep,lang2019pointpillars,zhou2018voxelnet,qi2018frustum,danzer20192d},
gesture recognition \cite{Wang2019},
semantic segmentation \cite{qi2017pointnet,qi2017pointnetplusplus,liu2019point},
and point set registration \cite{aoki2019pointnetlk,Sarode2019PCRNetPC,lu19,wang2019deep,gross2019alignnet,Wang2019PRNet}.
Core building block, which are common in these approaches, are point-wise feature embedding and permutation invariant feature aggregation by symmetric function.
Point-wise feature embedding is realized with the multi-layer perceptron (MLP) that processes input points; where the MLP itself consists of layers of matrix-vector product operation followed by nonlinearity that requires millions of dot products to process just one point.
The inputs points could be millions per second.
For example, a high-end LiDAR sensor obtains over 4 million points per second.
Therefore, it is very important to speed up the embedding step to realize real-time processing.
%  on the edge

To this end, we propose a novel framework, which we call \textit{LookUp Table Interpolation MLP (LUTI-MLP)} that can potentially replace any point feature embedding MLP used in PointNet-like architectures \cite{qi2017pointnet} to speed up the process.
At the time of testing, the LUTI-MLP computes the embedding using a linear combination of basis functions stored in a lookup table (LUT), which means the embedding no longer directly depends on MLP; rather, it depends on pre-computed basis functions, which results in the process being significantly more efficient than MLP embedding (\Sec\ref{subsec:embedding}). 
Furthermore, this approach enables us to very efficiently compute the analytical Jacobian wrt input point coordinate, which is another main contribution of this study, and one that could be used for real-time point-set registration/tracking (\Sec\ref{subsec:jacobian}).
Our contribution can be summarized as follows.

\prg{Fast Embedding.}
With the LUTI-MLP, the embedding computation at test time requires only looking up the nearest entities from the LUT that have been pre-computed for the irregularly arranged input space lattice $\mathbb{D}^3$, followed by interpolation (\Fig\ref{fig:overview}, middle, \textit{blue} line); it is much faster than MLP that involves computationally intensive matrix-vector product operation.

\prg{Fast Jacobian of Embedding.}
By minimizing the difference between the aggregated embedding features from pairs of point sets, one can compute a relative pose between them \cite{aoki2019pointnetlk}.
For this, the Jacobian of the features wrt global pose coordinate (Lie algebra $\mathfrak{se}(3)$) must be computed online at test time.
The local independence of the embeddings from the MLP wrt input points coordinate also provides efficient direct differential operation of the embedding wrt input in $\mathbb{R}^3$; that can be pulled back to the differential wrt $\mathfrak{se}(3)$ (\Fig\ref{fig:overview}, middle, \textit{red} line).
 
\prg{Experiments.}
With extensive evaluations on the ModelNet40 \cite{wu20153d}, we confirmed that the LUTI-MLP  embedding is about 100$\times$ faster than MLP embedding while achieving a comparable performance (proposed: $86.57$\%, original: $86.23$\%) using a very small table ($4^3$ lattice).
% $4^3$ lattice which is 2.3x smaller than that of embedding MLP requiring only 250 KB
The LUTI-MLP also provides significant speedup for the Jacobian computation wrt global pose: 12$\times$ for the approximate Jacobian and 860$\times$ for the analytical Jacobian.
The speedup of embedding and its Jacobian enables the state-of-the-art (SOTA) computational efficiency in several architectures and applications.
\section{Method}
\label{sec:method}
In this section, we describe LUTI-MLP, a novel point feature embedding architecture that significantly speeds up the embedding and its Jacobian computation at test time.
After preliminaries (\Sec\ref{subsec:preliminary}), we first review PointNet \cite{qi2017pointnet} (\Sec\ref{subsec:pointnet}), which enables the permutation invariant modeling of the points set, and then we formalize the problems under consideration (\Sec\ref{subsec:problem}).
Next, we introduce LUTI-MLP, a specifically designed irregularly tabulated MLP (\Sec\ref{subsec:luti_mlp}), which enables efficient linear computation of nonlinear embedding at test time (\Sec\ref{subsec:embedding}) and highly efficient Jacobian computation of the embedding function wrt global pose at test time (\Sec\ref{subsec:jacobian}).
Finally, the training methods are discussed (\Sec\ref{subsec:training}).

\subsection{Preliminaries}
\label{subsec:preliminary}
We consider a 3D geometric point set, $\mathbf{P}=\{\mathbf{p}_i |i = 1, ..., N\}$, where each point $\mathbf{p}_i$ is a vector of its 3D coordinate.
We model the relationship between the input point set and its output as
\begin{equation}
\mathbf{o}=f(\mathbf{P}),
\end{equation}
where output $\mathbf{o}$ could be the class label for the 3D object classification task, or it could be the point-wise label for the semantic segmentation task.
Function $f$ has to be invariant to permutations of its set members (e.g., the point cloud of a bunny is still a bunny, regardless of the order of each point in the set). 
% To realize such functions using neural networks, a permutation of input points would be a huge problem .
% A simple MLP or RNN \cite{Rumelhart:1986we} trained with randomly permuted sequences cannot scale to thousands or tens of thousands of input elements \cite{qi2017pointnet}. 

\subsection{Point-Feature Embedding by MLP}
\label{subsec:pointnet}
PointNet \cite{qi2017pointnet} (\Fig\ref{fig:overview}., top)  embeds each input point independently into high-dimensional feature space as
\begin{equation}
\label{eq:mlp_basic}
\mathbf{z}_i=\phi_{\text{MLP}}(\mathbf{p}_i),
\end{equation}
and then aggregates the set of embedded features $\mathbf{Z}:\{\mathbf{z}_1,...,\mathbf{z}_N\}$ to global feature $\mathbf{a}$ by the symmetric function $\max$ as
$\mathbf{a}=\max(\mathbf{Z})$,
where the nonlinear embedding function  $\phi_{\text{MLP}}:  \mathbb{R}^3 \rightarrow \mathbb{R}^K$ is  realized with MLP, and $\max: \mathbb{R}^K\times ...\times  \mathbb{R}^K \rightarrow \mathbb{R}^K$ operates along the point dimension. 
Then, global feature $\mathbf{a}$ is processed by function $g$ to compute the output as $\mathbf{o}=g(\mathbf{a})$.
Because of the symmetry of the aggregation function, the permutation of the input point set does not change the global feature $\mathbf{a}$, so output $\mathbf{o}$ does not change as a result.
Output function $g$ differs according to the type of task.
\prg{Object Classification.}
A global feature is fed directly to the classification MLP (mlp-c in \Fig \ref{fig:overview}) to output the scores for the candidate classes \cite{qi2017pointnet,qi2017pointnetplusplus,qi2018frustum,zhou2018voxelnet,lang2019pointpillars}.

\prg{Point-Wise Classification}
The concatenation of point-wise embedding features and a global feature is processed by another classification MLP to output the point-wise score for the candidate classes  \cite{qi2017pointnet,qi2017pointnetplusplus}.

\prg{Point Set Registration.}
The proximity of global features $|\mathbf{a}_\mathcal{S}-\mathbf{a}_\mathcal{T}|$, where  $\mathbf{a}_\mathcal{S}$ and $\mathbf{a}_\mathcal{T}$ are computed from the source and target point sets,  relates to the geometric proximity between the sets in the Euclidean input space; it is used to compute the geometric transformation between the sets \cite{aoki2019pointnetlk}. 

\subsection{Problem Statement}
\label{subsec:problem}
Our research goal is to derive an efficient method for computing nonlinear embedding $\mathbf{z}\in \mathbb{R}^K$ and its differential operation wrt global pose coordinates $\mathfrak{se}(3)$.
When processing 3D point clouds for real-time recognition, thousands or tens of thousands of input points must be processed dozens of times within a second.
This means the embedding function $\phi_{\text{MLP}}$ realized by the MLP must be evaluated millions of times per second.
This is quite difficult because the MLP consists of layers of matrix-vector product operations, which require millions of dot products per point input.
Thus, the speedup of the embedding is critical and has great practical importance.
Similarly,  the Jacobian of the global feature $\mathbf{a}$ wrt geometric transformation in $\mathfrak{se}(3)$ must be computed efficiently for point-set registration \cite{aoki2019pointnetlk} or tracking tasks.

\subsection{LUTI-MLP}
\label{subsec:luti_mlp}
By pre-computing the input-output relation of the embedding MLP of \eqref{eq:mlp_basic} on LUT, it would be possible to drastically speed up test time computation.
However, this is not feasible because of the memory footprint. 
For example, the LUT for $1024^3$ discretized inputs requires 4 terabytes of memory when $K=1024$\footnote{See \Supp{\ref{suppsec:memory}} for a more detailed discussion of the memory footprint.}.
% , which could not be stored even on high-end GPUs
Simply reducing the resolution induces an error, which is particularly severe when the discretization is coarse (\Sec\ref{subsec:analysis}). 

Instead of approximating a trained embedding MLP using LUT, we propose a novel end-to-end (E2E) trainable neural network framework called LUTI-MLP, which computes the embedding very efficiently by using the LUT structure while requiring a much smaller memory footprint than simple approximations by discretization.
LUTI-MLP computes the embedding as a weighted sum of the basis function $\phi_D(\mathbf{\bar{p}}^j)$, where the weight is computed using the spatial proximity between input $\mathbf{p}_i$ and discrete grid point $\mathbf{\bar{p}}^j_i$ on the lattice:
\begin{equation}
\label{eq:lutimlp_basic}
\mathbf{z}_i=\phi(\mathbf{p}_i)=\sum_j w^{(j)}(\mathbf{p}_i,\mathbf{\bar{p}}^{(j)}) \phi_{D}(\mathbf{\bar{p}}^{(j)}).
\end{equation}
The basis function $\phi_D(\mathbf{\bar{p}}^j)$ is computed from MLP using discrete inputs on lattice $\mathbb{D}^3$.
Notice that the basis functions $\phi_D(\bar{\mathbf{p}}^{(j)})$ are locally independent from input coordinate $\mathbf{p}_i$ in contrast to the direct dependance on $\phi_{\text{MLP}}$ in \eqref{eq:mlp_basic}.
At test time, this formulation leads to the drastic speedup of embedding and Jacobian computation by using pre-computed basis functions $\phi_D(\bar{\mathbf{p}}^{(j)})$ rather than evaluating the MLP.  
Note that the precise input locations are preserved regardless of the lattice resolution $D$ in the case of LUTI; contrarily, this is not the case for the LUT approximation.
Lattice resolution $D$ and weight function $w$ controls the expressiveness of the embedding function.

\subsection{Fast Embedding Computation at Testing}
\label{subsec:embedding}
At test time, the embedding feature $\mathbf{z}$ in \eqref{eq:lutimlp_basic} is computed as a linear combination of pre-computed basis function  $\phi_D(\bar{\mathbf{p}}^{(j)})$ stored in the LUT.
Memory lookup followed by interpolation is much cheaper than evaluating the MLP, which involves computationally intensive matrix-vector product operations.
We propose two types of interpolation for input in  $\mathbb{R}^3$: trilinear interpolation on uniform lattice ($\text{LUTI}_{\text{uni}}$), and trilinear interpolation on irregular lattice ($\text{LUTI}_{\text{irr}}$).
%%%%%%%%%%%%%%%%%%%%%%%%%%%%%%%%%%%%%
\prg{$\text{LUTI}_{\text{uni}}$.}
The continuous input point $\mathbf{p}_i$ is discretized to 8 neighborhood points, $\bar{\mathbf{p}}_i^{(000)},...,\bar{\mathbf{p}}_i^{(111)}$on the lattice, 
and the 8 pre-computed basis functions $\phi_D(\bar{\mathbf{p}}^{(j)}_i)$ corresponding to these discretized 8 input points are looked up.
Then, they are trilinearly interpolated using the Euclidean distances between the input and the neighborhood lattice as the weight, that is,
\begin{multline}
\label{eq:trilinear}
\mathbf{z}_i^{\text{uni}}
=\bar{\mathbf{z}}_i^{(000)}\bar{d_x}\bar{d_y}\bar{d_z}
+\bar{\mathbf{z}}_i^{(001)}\bar{d_x}\bar{d_y}\aar{d_z}
+\bar{\mathbf{z}}_i^{(010)}\bar{d_x}\aar{d_y}\bar{d_z}
+\bar{\mathbf{z}}_i^{(011)}\bar{d_x}\aar{d_y}\aar{d_z}\\
+\bar{\mathbf{z}}_i^{(100)}\aar{d_x}\bar{d_y}\bar{d_z}
+\bar{\mathbf{z}}_i^{(101)}\aar{d_x}\bar{d_y}\aar{d_z}
+\bar{\mathbf{z}}_i^{(110)}\aar{d_x}\aar{d_y}\bar{d_z}
+\bar{\mathbf{z}}_i^{(111)}\aar{d_x}\aar{d_y}\aar{d_z},
\end{multline}
where $d=p \rceil-p$ and $\bar{d}=p-p\rfloor$ are the Euclidean distance to the neighbors on the lattice (\Fig\ref{fig:overview}., middle). 
This interpolation may suffer from performance degradation depending on the task when the lattice is coarse. 
% The precise input point locations are preserved on each embedding vector by the interpolation.

\begin{figure}[!h]
\begin{center}
\includegraphics[width=0.97\columnwidth,trim={0 0.1cm 0 0.02cm},clip]{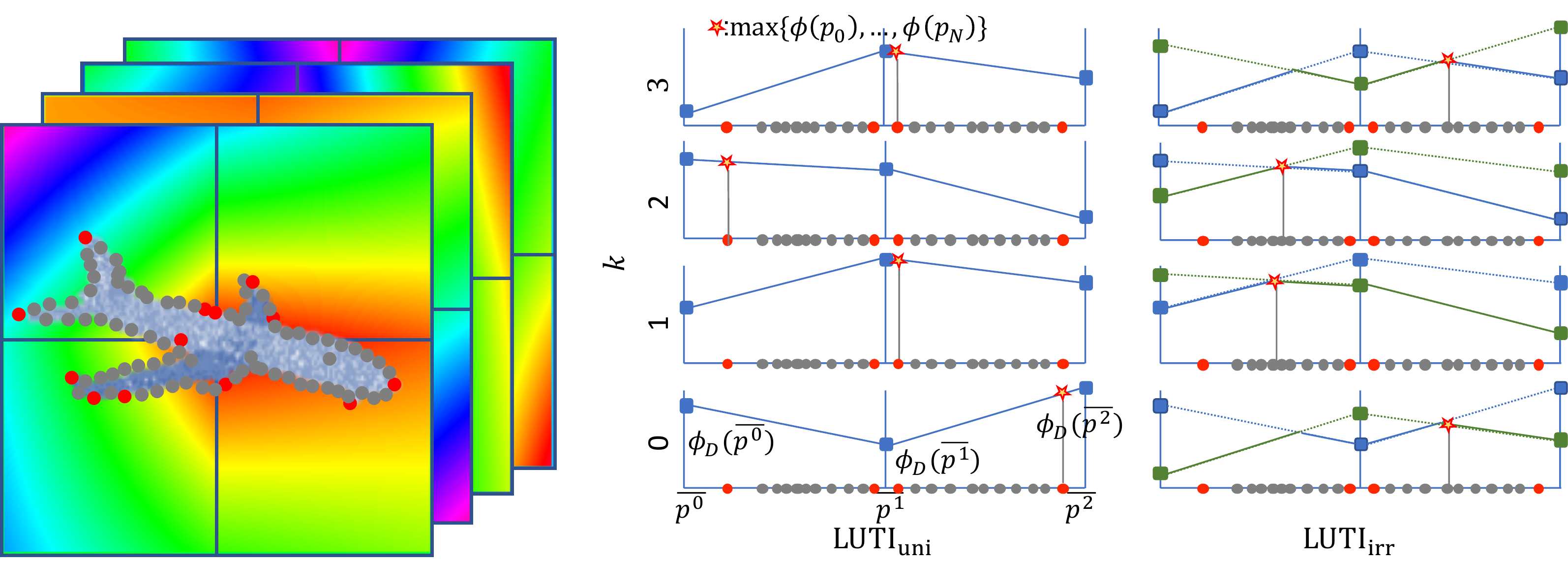}
\end{center}{}
\caption{
\textbf{Interpolation of basis functions on regular/irregular grid.}
% Problems of interpolation on regular grid and proposed efficient interpolation on irregular grid are illustrated in the case of $K=4$. 
Left: Trilinear interpolation on a uniform grid limits the number of active input points $\mathbf{P}_{\text{a}}$  (the gray points can not be $\mathbf{P}_{\text{a}}$ across the channel), because $\arg\max_{\mathbf{p}}{\phi(\mathbf{p}})$ is fixed on the same uniformly allied grid points across channel.
Middle: This is illustrated in the case of 1D linear interpolation ($K=4$), where the points in between the red points have no chance of being the maximum ($\textcolor{red}{\star}$) regardless of the learned basis.
Right: The minimum of two interpolated values (solid line) can be at a maximum at any point depending on the learned  basis function. 
This is substantially the same as learning the irregular lattice location.
\label{fig:intp}
}
\end{figure}

%%%%%%%%%%%%%%%%%%%%%%%%%%%%%%%%%%%%%
\prg{$\text{LUTI}_{\text{irr}}$.}
Consider active input points $\mathbf{P}_{\text{a}}$ that contribute to global vector $\mathbf{a} = \max\{\mathbf{z}_1,...,\mathbf{z}_N\}$.
In the case of MLP embedding, $\arg\max_{\mathbf{p}}{\phi_{MLP}(\mathbf{p}})$ can vary per channel.
This means, at most, $K$ input points can be $\mathbf{P}_{\text{a}}$.
However, in the case of $\text{LUTI}_{\text{uni}}$, few points closest (in term of interpolated value) to the grid points can be $\mathbf{P}_{\text{a}}$ (\Fig\ref{fig:intp}., left and middle).
In other words, $|\mathbf{P}_{\text{a}}|$ is restricted by lattice resolution $D$ rather than $K$, and most of the input points cannot contribute to $\mathbf{a}$ when $D$ is coarse ($D^3\ll K$). 
It could negatively affect the performance due to the loss of detail information.
For example, when 2,048 points are fed to the network with $D=2$ and $K=2048$, approximately 50 points that amount to only a small contribution of about 2\% to $\mathbf{a}$.
% (in the case where a segmentation network is used, as in \Sec\ref{subsec:exp_app})
This can be solved by changing the grid locations for each channel.
Although it may be possible to train such per channel lattice locations in addition to the basis function and apply trilinear interpolation independently for each channel, it is computationally demanding and spoils the efficiency of LUTI-MLP.
Instead, we propose a simple yet efficient method that is substantially equivalent to varying the lattice location channel by channel using a minimum of two interpolated vectors computed using uniform lattice as, 
\begin{equation}
\label{eq:ccv}
\mathbf{z}_i^{\text{irr}} = \min\{\mathbf{z}_i^{\text{uni}},\Gamma(\mathbf{z}_i^{\text{uni}})\},
\end{equation}
where $\Gamma$ reverses the input along the channel dimension. 
The interpolated value can be maximized at any point; thus, $|\mathbf{P}_{\text{a}}|$ is no longer restricted by the lattice resolution (\Fig\ref{fig:intp}, right).
The effects of the interpolation method on the performance are discussed in \Sec\ref{subsec:exp_app} and \Sec\ref{subsec:analysis} and learned embedding spaces are visually compared in \Fig\ref{fig:vis_LUTI}.
% In addition, learned embedding space using this irregular interpolation are visually compare with regular one in \Fig\ref{fig:vis_LUTI}.
% \footnote{For more detailed discussion  about irregular trilinear interpolation, see  \Supp{\ref{suppsec:irregular}}}

% However,\eqref{eq:trilinear} limit the the numbers of maximum contribution to $\text{D}^3$; 
% It is obvious from \eqref{eq:trilinear} that the for any embedding channel $k$, interpolated value takes maximum at lattice location.
% This means, only input points that is close to the lattice has chance to be maximum, and points that are not has no chance regardless of the learned basis function.
% We experimentally confirmed that even a very coarse lattice resolution ($\text{D}=4$) yields excellent performance comparable with MLP while achieving about 100x speedup.

\subsection{Fast Jacobian Computation wrt Pose at Testing}
\label{subsec:jacobian}
Using the formulation of LUTI-MLP in \eqref{eq:lutimlp_basic}, the Jacobian of each element of embedded feature $\mathbf{z}$ wrt to the input coordinates  $(x,y,z) \in \mathbb{R}^3$ in Euclidean space is computed quite efficiently.
In the case of $\text{LUTI}_{\text{uni}}$, it is derived from \eqref{eq:trilinear}:

\begin{multline}
\label{eq:jacobian}
\frac{\partial \phi(\mathbf{p})_i}{\partial x}
=(\bar{\mathbf{z}}_i^{(000)}-\bar{\mathbf{z}}_i^{(100)})(\bar{d_y}\bar{d_z})
+(\bar{\mathbf{z}}_i^{(010)}-\bar{\mathbf{z}}_i^{(110)})(\aar{d_y}\bar{d_z})\\
+(\bar{\mathbf{z}}_i^{(001)}-\bar{\mathbf{z}}_i^{(101)})(\bar{d_y}\aar{d_z})
+(\bar{\mathbf{z}}_i^{(011)}-\bar{\mathbf{z}}_i^{(111)})(\aar{d_y}\aar{d_z}),
\end{multline}
which is a bilinear interpolation of the difference of the  basis functions $\phi_D(\bar{\mathbf{p}}^{(j)})$ that is also pre-computable.
Jacobian wrt $y$ and $z$ are computed similarly.
In the case of $\text{LUTI}_{\text{irr}}$, it is derived from \eqref{eq:ccv}.
This Jacobian wrt $\mathbb{R}^3$ is easily pulled back to the Jacobian wrt geometric coordinates $\boldsymbol{\xi}$  in  $\mathfrak{se}(3)$ as $\partial \mathbf{p}/\partial\boldsymbol{\xi} = [[\mathbf{p}]_{\times}, \mathbf{I}]$, 
% \begin{align}
% \label{eq:dpdxi}
% \frac{\partial \mathbf{p}}{\partial\boldsymbol{\xi}}=
% \left( {\begin{array}{*{20}{c}}
%   0&{ - z}&{ y} \\ 
%   { z}&0&{-x} \\ 
%   {-y}&x&0 
% \end{array}\begin{array}{*{20}{c}}
%   1&0&0 \\ 
%   0&1&0 \\ 
%   0&0&1
% \end{array}} \right).
% \end{align}
% \begin{align}
% \label{eq:dpdxi}
% \frac{\partial \mathbf{p}}{\partial\boldsymbol{\xi}}=
% \left
% [[\mathbf{p}]_{\times}, \mathbf{I}]
% \right.
% \end{align}
and the analytical Jacobian of embedded feature $\mathbf{z}$ wrt $\boldsymbol{\xi}$ is given as

\begin{equation}
\label{eq:dadxi}
\frac{\partial \phi(\mathbf{p})}{\partial \boldsymbol{\xi}}=\frac{\partial  \phi(\mathbf{p})}{\partial \mathbf{p}}\frac{\partial \mathbf{p}}{\partial \boldsymbol{\xi}}.
\end{equation}

\prg{Application to Point Set Registration.}
PointNetLK \cite{aoki2019pointnetlk} proposed the use of global features for 3D point cloud registration.
Unlike iterative closest point (ICP) that is commonly used for point set registration, this approach requires no costly computation of point correspondences \cite{rusinkiewicz2001efficient}, which provides advantages in terms of computational efficiency.
The core building block of this type of architecture is computation of Jacobian wrt global pose coordinates $\mathfrak{se}(3)$; computation of Gauss–Newton updates using the Jacobian  dominate the overall processing time.
LUTI-MLP Jacobian  could further boost the speed for real-time point-cloud registration/tracking.
Proximity of the source $\mathbf{P}_{\mathcal{S}}$ and target  $\mathbf{P}_{\mathcal{T}}$ point set is measured as
\begin{equation}
\label{eq:invj}
\mathbf{r}=[\max(\phi(\mathbf{G\cdot P}_{\mathcal{S}})-\max(\phi(\mathbf{P}_{\mathcal{T}}))].
\end{equation}
The geometric transformation $\mathbf{G}\in SE(3)$ between sets is estimated as a minimizer of the residual  $\mathbf{r}$ wrt $\mathbf{G}$. 
Jacobian $\mathbf{J}$ for the aggregated embedded feature $\max(\phi(\mathbf{P}_{\mathcal{T}}))$ wrt $\boldsymbol{\xi}\in\mathfrak{se}(3)$ is used in inverse compositional formulation \cite{baker2004lucas} as
$\boldsymbol{\Delta\xi}=\mathbf{J}^{\dagger}\mathbf{r},$
where $\mathbf{J}^{\dagger}$ is the Moore-Penrose inverse of  $\mathbf{J}$, which is computed only once for a given target point set  $\mathbf{P}_{\mathcal{T}}$, and residual $\mathbf{r}$ is computed iteratively using $\mathbf{G}$  until convergence, where it is updated as $\Delta \mathbf{G}=\exp \left(\sum_{k=1,...,6} \Delta\boldsymbol{\xi}_{k} \mathbf{T}_{k}\right)$.
The acceleration of LUTI-MLP can be used to compute $\phi(\mathbf{G\cdot P}_{\mathcal{S}})$ in the iterative residual computation, and it can also be used for Jacobin computation (in two ways), which are discussed below.

\prg{Approximate Jacobian.}
Given the difficulty of computing the analytical Jacobian of the MLP, the authors in \cite{aoki2019pointnetlk} used the finite difference method (FDM) to approximate the Jacobian as
\begin{equation}
\label{eq:approx_lk}
\mathbf{J}_{k}=\frac{\max(\phi(\exp (-t_{k} \mathbf{T}_{k})\cdot \mathbf{P}_{\mathcal{T}}))-\max(\phi(\mathbf{P}_{\mathcal{T}}))}{t_{k}},
\end{equation}
where $t_k$ is the small perturbation in the $k$-th element of the $\mathfrak{se}(3)$ parameter.
We can replace embeddings $\phi$ realized as the MLP with the proposed LUTI-MLP to speed up the computation of this FDM.

\prg{Analytical Jacobian.}
Instead of the numerical differentiation, \eqref{eq:dadxi} gives an efficient analytical Jacobian of global feature $\mathbf{a}$ wrt  $\boldsymbol{\xi}$ as
\begin{align}
\label{eq:canonic_lk}
    \mathbf{J}&=\frac{\partial \mathbf{p}}{\partial \boldsymbol{\xi}}\frac{\partial}{\partial \mathbf{p}}[\max(\phi(\mathbf{G}^{-1} \cdot \mathbf{P}_{\mathcal{T}}))].
\end{align}
The analytical Jacobian has practical advantages that it has no hyper-parameter $t_k$ in \eqref{eq:approx_lk} (a small perturbation) for computing the finite difference whose appropriate value may differ depending on the network architecture or data to process. 
Note that the computation of the analytical Jacobian of MLP wrt $\mathbf{p}$ and $\boldsymbol{\xi}$ is quite complicated, requiring the partial derivative computation of each element of embedding feature $\mathbf{z}$ wrt nonlinear MLP.
It is computed by looping the backpropagation from each element of $\mathbf{z}$ for $K$ times.

\subsection{Training LUTI-MLP}
\label{subsec:training}
LUTI-MLP is differentiable and can be trained E2E (\Fig\ref{fig:overview}, bottom) because $\phi_D$ and the interpolation of \eqref{eq:lutimlp_basic}  are differentiable. 
The basis function $\phi_D(\bar{\mathbf{p}}^{(j)})$, which is actually an MLP, is trained using standard back-propagation using the errors from the upper layers.
The back-propagated errors on the interpolated embedding $\mathbf{z}_i$  are distributed to the basis function $\phi_D(\bar{\mathbf{p}}^{(j)}_i)$  on the lattice according to the weight $w^{(j)}(\mathbf{p}_i,\mathbf{\bar{p}}^{(j)})$.
Because the same interpolation is used at the training and test time, the embedding $\mathbf{z}$ computed using pre-computed $\phi_D$ on  the LUT at test time is exactly the same as the one computed using the MLP during training.
As discussed in \Sec\ref{sec:exp}, the approximation of trained embedding MLP using trilinear interpolation yields a poor performance;
we experimentally revealed that E2E training is key to achieve an excellent performance comparable to MLP embedding, especially when the lattices are coarse.

\section{Experiments}
\label{sec:exp}
This section report the intensive experimental results to evaluate the effectiveness of LUTI-MLP.
Our main focus is evaluating the distinct idea of LUTI-MLP, feature embedding by a linear combination of non-linear basis vectors on an irregular lattice, in its most basic setup (model/data).
Therefore, the application of LUTI-MLP to more complicated recent models \cite{qi2017pointnetplusplus,thomas2019kpconv,dgcnn,wang2018local,landrieu2018large} and its evaluation on real-world datasets is left for the future work.
First, we show the speed-up gain by using LUTI-MLP for embedding and Jacobian calculations (\Sec \ref{subsec:exp_speed}). 
Second, we show the applicability of the LUTI-MLP to several architectures, such as ones designed for object classification, point-wise classification, and point-cloud registration (\Sec \ref{subsec:exp_app}).
Third, we provide an intensive architectural analysis to find the key components for the good performance of the LUTI-MLP (\Sec \ref{subsec:analysis}). 
Fourth, we visualize what the embedding network with different discretization learns (\Sec \ref{subsec:exp_vis}).

\subsection{Computational Speed Analysis}
\label{subsec:exp_speed}
We evaluated the speedup gain by using the LUTI-MLP over the MLP for embedding and Jacobian computation.
In this subsection, we focus our discussion on embedding and Jacobian computation parts that are common to a range of architectures, and the overall speedup gain for a specific architecture is discussed in the subsequent subsection.
For the speed benchmarking, we implemented the Nvidia CUDA kernel for trilinear interpolation (embedding)  and bilinear interpolation (Jacobian).
Batch normalization \cite{ioffe2015batch} used in the MLP, and it was integrated into the matrix-vector product layer for a fair comparison. 
% Cudnn acceleration were enabled to  the MLP.
For both experiments, we report the average wall-clock time evaluated on Nvidia V100 GPU to process 1,024 randomly generated points.

\prg{Embedding.}
\Tab\ref{table:speed}  compares the computational complexity of embedding between the MLP and the LUTI-MLP at test time.
The speedup gain from LUTI acceleration is about 100$\times$ (at $D=4$, yielding similar accuracy as with MLP, see \Fig\ref{fig:effect_D}) compared with the highly optimized cudnn implementation of the MLP.
Although the number of lookups and the linear interpolation of LUTI are constant across different lattice resolution (except $D=2$, where no look-up operation is required),  the computational time at $D=4$ is 35\% faster than  at $D=64$.
It is because the memory access tends to be more efficient for a small LUT.
Note that the MLP version slows down as the number of layers or intermediate channel increases, while LUTI is invariant to the architecture.
\begin{table}
\vspace{-8mm}%Put here to reduce too much white space after your table 
\caption{\label{table:speed}
\textbf{Time complexity of feature embedding.}
We report the latency ($\mu$s) for embedding 1,024 points using MLP and LUTI with different lattice resolution $D$. 
LUTI$_{\text{uni}}$ uses interpolation on a regular grid of \eqref{eq:trilinear}, and LUTI$_{\text{irr}}$ uses interpolation on an irregular grid of \eqref{eq:ccv}.
The MLP is the same as that of PointNet, and the times (accelerated by cudnn) are reported in parentheses.
Both of the proposed LUTI embeddings are much faster than MLP.
% For LUTI, the times for the different interpolation methods () are reported, 
% accelerated by cudnn.
% The benchmark was done with and without cudnn acceleration for MLP.
% Despite the number of lookup and linear interpolation of LUTI are the same across different $D$, its computation time increase as $D$ increase for small $D$ and became almost constant for $D>32$.
% It is because, the memory access tend to take more time for larger LUT.
}

\begin{center}
\begin{tabular}{c|c|cccccc|cccccc}
\multicolumn{1}{c}{} & \multicolumn{1}{c}{MLP}  
& \multicolumn{6}{c}{$\text{LUTI}_{\text{uni}}$}
& \multicolumn{6}{c}{$\text{LUTI}_{\text{irr}}$}

\tabularnewline
% \hdashline
$D$ & -  &  
64 & 32 & 16 & 8 & 4 & 2 &
64 & 32 & 16 & 8 & 4 & 2
\tabularnewline
\hline 
Latency ($\mu$s)& 4330 (4350) &   
71.8 & 71.6 & 69.8 & 53.6 & 46.8 & 34.2 &
74.9 & 75.5 & 73.7 & 57.0 & 50.6 & 37.3
\tabularnewline
\end{tabular}
\end{center}
\vspace{-8mm}%Put here to reduce too much white space after your table 
\end{table}

\prg{Jacobian.}
% https://context.reverso.net/translation/english-japanese/the+following+table+compares
\Tab\ref{table:speed_jac} (Jac.) compares the test time complexity of different types of Jacobian computation using the MLP and $\text{LUTI}_{\text{irr}}$-MLP: approximation using FDM of MLP \cite{aoki2019pointnetlk}, approximation using FDM of LUTI-MLP, analytical differentiation of MLP, and analytical differentiation of LUTI-MLP.
The computation of the approximate Jacobian includes the first term of the numerator of \eqref{eq:approx_lk}, which is the warping of point clouds, followed by embedding and aggregation by $\max$ function.
The computation of analytical Jacobian using the LUTI includes bilinear interpolation of \eqref{eq:jacobian} and the index copy operation using $\argmax$.
The computation of the analytical Jacobian of the MLP requires $K$ times backpropagation\footnote{We use the parallelization technique disclosed in https://gist.github.com/sbarratt (torch\_jacobian.py) to accelerate the multiple backward computations.}, and the same index operation is followed. 
The LUTI accelerates the computation of the approximate Jacobian for 12$\times$ and the analytical Jacobian for 860$\times$ compared to MLP with cudnn. 
% https://www.tablesgenerator.com
\begin{table}
\vspace{-8mm}%Put here to reduce too much white space after your table 
\caption{\label{table:speed_jac}
\textbf{Time complexity of  Jacobian computation and pose update.}
We compare the computational time (ms) of four different types of Jacobians (Jac.) and the corresponding time for the whole pose estimation  (All):
FDM of MLP \cite{aoki2019pointnetlk} (FDM-MLP), FDM of LUTI$_{\text{irr}}$-MLP (FDM LUTI), analytical  Jacobian of MLP (Analyt.-MLP), and analytical  Jacobian of LUTI$_{\text{irr}}$-MLP (Analyt.-LUTI).
$D=8$ is used for LUTI$_{\text{irr}}$-MLP.
For MLP, timing accelerated by cudnn are reported in parentheses
}
\begin{center}
\begin{tabular}{c|cc||c|cc}
\multicolumn{1}{c}{ Algorithm }  &  Jac.  & All  & \multicolumn{1}{c}{ Algorithm }  &  Jac.  & All  \\
\hline
FDM-MLP \cite{aoki2019pointnetlk} &  5.34 (4.86) & 93.3 (92.5) & Analyt.-MLP & 786 (437) & 874 (525) \\
FDM-LUTI (ours)   & 0.401 & 2.47 & Analyt.-LUTI (ours)  & 0.503 & 2.58 \\
\end{tabular}
\end{center}
\vspace{-8mm}%Put here to reduce too much white space after your table 
\end{table}

% Our LUTI accelerates the computation of FDM version of Jacobian ($12\times$), canonical Jacobian ($860\times$).
% pose estimation using approximate Jacobian ($37\times$), and pose estimation using canonical Jacobian ($200\times$) respectively compared with MLP with cudnn.  

% $\partial \mathbf{a}/\partial \mathbf{\boldsymbol{\xi}}$ 
%  $\max(\phi\left(\exp \left(-t_{k} \mathbf{T}_{k}\right) \cdot \mathbf{P}_{\mathcal{T}}\right)$,
% , which is assumed to be pre-computed.

\subsection{Applications}
\label{subsec:exp_app}
We demonstrate applicability of LUTI-MLP to several architectures, such as 3D object classification, 3D object part segmentation, and 3D point-set registration. 
% We evaluated our PointNet and PointNetLK with LUTI-MLP against the original network using MLP for embedding.
In summary, we observe an overall speedup by LUTI-MLP embedding without performance deterioration for different types of architectures and applications.
For these experiments, we  followed the original description for the basic network architecture and training procedure of PointNet \cite{qi2017pointnet} for classification and segmentation, and PointNetLK \cite{aoki2019pointnetlk} for point-set registration.
The only difference from the original description is the use of  LUTI-MLP instead of MLP for embedding.
No other modification, such as hyper-parameter tuning, was performed.
For all experiments, we used the PyTorch \cite{paszke2017automatic} library for training and testing.
\prg{3D Object Classification.}
Given a 3D point cloud, the task is to assign an object class label for the point set. 
% The results are summarized in \Fig \ref{fig:effect_D}.
Ours (\Fig \ref{fig:effect_D}. LUTI$_{\text{irr}}$-MLP E2E with $D=4$) achieves 86.57\%, which slightly outperforms the original network using MLP embedding (86.23\% shown by the dashed line), while achieving 34$\times$ speedup of overall classification process.
Interpolation on regular grid (LUTI$_{\text{uni}}$) showed slight performance deterioration when lattice are very coarse, while LUTI$_{\text{irr}}$ performed equally well even for the coarsest lattice.
The evaluation is conducted on the ModelNet40 \cite{wu20153d} which contains 12,311 models from 40 categories.
% , split into $9,843$ for training and $2,468$ for testing.
We randomly sampled $1,024$ points and normalized them into a unit sphere. 
% During training, we followed the same data argumentation protocol of the PointNet \cite{qi2017pointnet},  augment the point cloud on-the-fly by randomly rotating the object along the up axis, and jitter the position of each point by Gaussian noise with zero mean and 0.02 standard deviation.
% We also provide the detailed architectural design analysis using the this task in \Sec\ref{subsec:analysis}.

\prg{3D Object Part Segmentation.}
Given a 3D point cloud, the task is to assign a part category label to each point\footnote{Point-wise classification network used in this task is shown in \Supp{\ref{suppsec:seg}}.}. 
The evaluation metric is mIoU on points \cite{qi2017pointnet}.
For this point-wise classification task, architecture from \cite{liu2019point} utilizing both point and voxel representation is SOTA in terms of accuracy and latency.
Our network with LUTI$_{\textbf{irr}}$-MLP showed comparable performance to PointNet \cite{qi2017pointnet} and PVCNN (0.25$\times \text{C}$ that is fastest variant on this task) \cite{liu2019point} (\Tab\ref{table:shapenet_digest}) while achieving 
$1.9\times$ and $1.3\times$ overall speedup, respectively \footnote{The original PVCNN used 6-dimensional vectors (XYZ+normals) as input, but the results in \Tab\ref{table:shapenet_digest}  use 3-dimensional vectors (XYZ) in order to align the experimental conditions, and this would be the main reason for the difference of performance.}.
The speedup gain of this type of architecture is rather moderate compared to other two; it is because the point-wise classification after the embedding accounts for a larger percentage of the entire processing.
Unlike the previous application, LUTI$_{\text{uni}}$ showed large performance drop in this task when lattice are very coarse, while LUTI$_{\text{irr}}$ performed equally well even for the coarsest lattice.
The evaluation is conducted on the ShapeNet \cite{shapenet2015}, which contains 16,881 models from $16$ categories.
We randomly sampled $2,048$ points and normalized them into a unit sphere. 
% Local feature $\mathbf{Z}^{feat}$ is also used for this network; the table size for these experiments is $ D^3\times$(64+128+128+512+2048).
% The results are summarized in \Tab\ref{table:shapenet_digest}.
\begin{table}
\vspace{-8mm}%Put here to reduce too much white space after your table 
\caption{\label{table:shapenet_digest}
$\textbf{Parts segmentation accuracy for the ShapeNet part data set.}$
Metric is mIoU(\%) on points.
The network using LUTI$_{\textbf{irr}}$-MLP embedding performs as well as those using the MLP embedding \cite{qi2017pointnet} and PVCNN \cite{liu2019point}
}

\begin{center}
\begin{adjustbox}{width=1\columnwidth}
\begin{tabular}{cc|c|cccccccccccccccc}
\hline
& & mIoU & aero & bag & cap & car & chair & \begin{tabular}{c}ear\\phone\end{tabular} & guitar & knife & lamp & laptop & motor & mug & pistol & rocket & \begin{tabular}{c}skate\\board\end{tabular} & table \\ \hline
\#shapes & \text{D} & & 2690 & 76 & 55 & 898 & 3758 & 69 & 787 & 392 & 1547 & 451 & 202 & 184 & 283 & 66 & 152 & 5271 \\ \hline
$\text{MLP}$ \cite{qi2017pointnet} & - & 
83.0 & 82.7 & 77.9 & 74.9 & 75.4 & 89.1 & 66.0 & 91.3 & 85.4 & 80.3 & 94.7 & 66.7 & 92.2 & 79.9 & 50.3 & 71.1 & 80.4\\
\hdashline
$\text{PVCNN}$ \cite{liu2019point}& - & 
82.8 & 80.5 & 80.8 & 83.1 & 76.1 & 89.3 & 72.8 & 90.8 & 85.2 & 82.0 & 95.1 & 67.0 & 92.9 & 81.1 & 56.9 & 72.2 & 79.1 \\
\hdashline
$\text{LUTI}_{\text{uni}}$ & 2 &  
71.7 & 64.4 & 70.7 & 72.6 & 63.6 & 79.0 & 63.9 & 84.7 & 77.3 & 62.6 & 91.3 & 51.9 & 89.8 & 65.6 & 50.1 & 65.6 & 72.6\\
$\text{LUTI}_{\text{uni}}$ & 4 &  
77.0 & 74.2 & 71.3 & 69.9 & 66.2 & 83.0 & 61.2 & 86.9 & 79.6 & 72.4 & 92.4 & 60.0 & 88.3 & 74.2 & 47.8 & 69.7 & 76.3\\
$\text{LUTI}_{\text{uni}}$ & 8 &  
81.6 & 81.7 & 72.2 & 80.1 & 72.6 & 87.5 & 61.9 & 89.8 & 83.4 & 78.0 & 94.4 & 62.7 & 92.8 & 78.8 & 55.1 & 72.0 & 79.8\\
$\text{LUTI}_{\text{uni}}$ & 16 & 
83.1 & 82.7 & 74.7 & 79.0 & 74.0 & 89.3 & 70.3 & 91.2 & 85.9 & 79.4 & 94.6 & 66.3 & 91.2 & 82.6 & 49.5 & 73.4 & 81.0 \\
\hdashline
$\text{LUTI}_{\text{irr}}$ & 2 & 
82.9 & 81.8 & 75.5 & 82.5 & 72.7 & 89.0 & 71.5 & 90.5 & 83.9 & 79.8 & 95.3 & 65.0 & 92.9 & 82.4 & 53.1 & 74.2 & 81.1\\
$\text{LUTI}_{\text{irr}}$ & 4 &  
83.3 & 82.0 & 75.0 & 80.5 & 75.4 & 89.0 & 72.4 & 91.0 & 84.5 & 80.6 & 95.0 & 66.7 & 92.6 & 82.0 & 50.1 & 73.6 & 81.4\\
% 83.2 & 82.6 & 77.1 & 79.6 & 75.6 & 89.2 & 66.7 & 91.0 & 85.4 & 80.1 & 94.8 & 65.0 & 93.1 & 81.0 & 50.9 & 72.8 & 80.9\\
$\text{LUTI}_{\text{irr}}$ & 8 &  
83.1 & 82.7 & 74.7 & 79.0 & 74.0 & 89.3 & 70.3 & 91.2 & 85.9 & 79.4 & 94.6 & 66.3 & 91.2 & 82.6 & 49.5 & 73.4 & 81.0\\
$\text{LUTI}_{\text{irr}}$ & 16 &  
83.2 & 82.7 & 79.4 & 82.1 & 75.0 & 88.9 & 68.7 & 91.3 & 85.1 & 80.2 & 94.9 & 66.0 & 92.8 & 81.8 & 51.2 & 74.5 & 81.0 \\
\hline
\end{tabular}
\end{adjustbox}
\end{center}
\vspace{-8mm}%Put here to reduce too much white space after your table 
\end{table}
% about $1.9\times$ and $1.3\times$ speedup for overall processing respectively

\prg{3D Point Cloud Registration.}
\label{subsubsec:PTLK}
Given the source ($\mathbf{P}_{\mathcal{S}}$) and target  ($\mathbf{P}_{\mathcal{T}}$) 3D point clouds, the task is to estimate the 3D geometric transformation in $SE(3)$ between the sets.
In these experiments, we followed the procedure of PointNetLK \cite{aoki2019pointnetlk} and compared the registration accuracy after 10 iterations of optimization and computation time for the whole pose estimation process.
LUTI is used to accelerate iterative residual computation \eqref{eq:invj} and for Jacobian computation, either approximately \eqref{eq:approx_lk} or analytically \eqref{eq:canonic_lk}, as discussed in \Sec\ref{sec:method}.
Original PointNetLK is compared with one using the $\text{LUTI}_{\text{irr}}$-MLP ($D$=8) for approximate (FDM LUTI) and analytical (Analiyt. LUTI) Jacobian computation.
\Tab\ref{table:speed_jac} (All) compares the computational complexity, and both variants using the LUTI embedding show similar performance as the original network using the MLP embedding while realizing a overall registration process that is about 37x faster\footnote{The network and evaluation are shown in \Supp{\ref{suppsec:pnlk}}}.
Furthermore, we observed faster convergence in LUTI variant than MLP, it may attributed by the smoother embedding space of it (\Fig\ref{fig:vis_emb_LUT_irr}).
% The over all computational time for pose estimation, includes $20$ iterations of the residual evaluation of \eqref{eq:invj}, are also reported.
% Both variants using LUTI-MLP (one use approximate and the other use analytical Jacobian) achieved accuracy comparable to that of the original PointNetLK using MLP while achieving orders of magnitude speedup (\Tab\ref{table:speed_jac}).

\subsection{Architecture Design Analysis}
\label{subsec:analysis}
\begin{SCfigure}
\caption{
\label{fig:effect_D}
% \addvbuffer[5ex 0ex]{
\textbf{Architecture design analysis.}
This figure shows the accuracy of the object classification task on ModelNet40.
The dashed line represents the baseline (86.23\%) using the MLP embedding.
$\text{LUTI}_{\text{irr}}$-MLP trained E2E performed better than those using the MLP (PointNet) across all range of D while achieving significant speedup
}
\includegraphics[width=0.52\columnwidth]{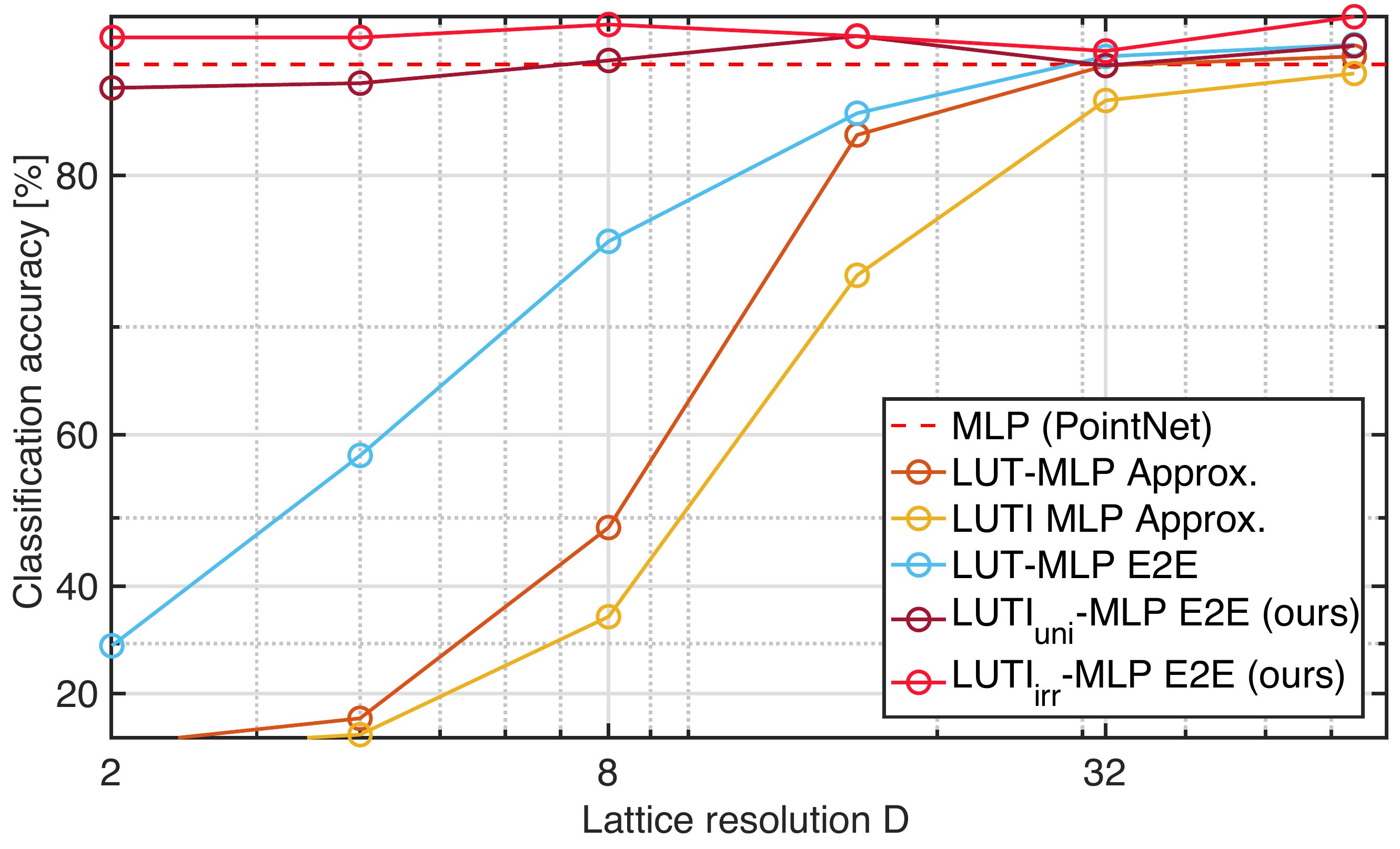}
\end{SCfigure}

We conducted an intensive ablation study\footnote{More analysis, e.g., use of T-Net \cite{qi2017pointnet} (\Supp{\ref{suppsec:tnet}}) and direct LUT(I) training without MLP are on \Supp{\ref{suppsec:detail_analisys}}.} to reveal the key components affecting the performance of the proposed architecture.
This evaluation used the same object classification task discussed in \Sec\ref{subsec:exp_app}, and the results are summarized in \Fig \ref{fig:effect_D}, showing the following:
% with varied lattice resolutions
\begin{itemize}
    \item The naïve approximation of the pre-trained model using LUT-storing shows good accuracy when fine  lattice is used (LUT(I)-MLP Approx., $D$$\gtrsim$64) although it is memory demanding and slower than the coarse one\footnote{The memory requirement is $\propto\! D^M$ (\Supp{\ref{suppsec:memory}}, \Tab\ref{table:memory}).}.
    \item E2E training of LUTI with irregular interpolation is crucial for achieving good accuracy with a coarse lattice ($\text{LUTI}_{\text{irr}}$-MLP E2E); this provides a promising scheme that simultaneously yields high speed, a feasible memory footprint, and good accuracy under realistic settings. 
\end{itemize}{}
% it quickly becomes infeasible when other modalities, such as surface-normal, are concatenated
% A finer lattice results in an infeasible memory footprint for storing the table (\Sec\ref{sec:method}). 

% We compared the proposed method against 5 architectures.
% T-Net \cite{qi2017pointnet}  to analyticalize the input point set by affine transformation can also be accelerated by LUTI-MLP with the slight modification\footnote{See \Supp{\ref{suppsec:tnet}} for detailed network architecture and additional results with FT}.

\vspace{-4mm}%Put here to reduce too much white space after your table 
% \begin{figure}[!h]
\vspace{-4mm}%Put here to reduce too much white space after your table 

% See the main text for the explanation of each algorithm.
% The metric is overall classification accuracy.
% }
% \vspace{1cm}
% \includegraphics[width=0.5\columnwidth,trim={12cm 0 0 0},clip]{fig/effect_D_.eps}
% \caption{
% \label{fig:effect_D}
% \textbf{Effects of table size and architectural design for the object classification performance on the Model-Net40 test set.}
% Lines without marker represents the results from baseline PointNet without and with T-Net for IT.
% The metric is overall classification accuracy.
% Our PointNet with LUTI-MLP shows comparable or slightly better results against naive PointNet across all range of $\text{D}$ while achieving significant speedup.
% See the main text for the explanation of each algorithm.
% }
% \end{figure}
% Performance comparison between several architectures and different discretization $\text{D}$ is reported.

\prg{MLP.}
MLP is a re-implementation of the PointNet \cite{qi2017pointnet}.
%  and found similar accuracy to that which was found  in the original paper.
The model was trained for a total of  $200$ epochs using the same scheduling as described in \cite{qi2017pointnet}; the other trained variants used the same scheduling.

\prg{LUT(I)-MLP Approx.}
These variants approximate the trained MLP of vanilla PointNet using the LUT or $\text{LUTI}_{\text{uni}}$.
LUT-MLP Approx. simply discretized input points to $\mathbb{D}^3$.
LUTI-MLP Approx. used the interpolation of \eqref{eq:trilinear}, but no E2E training was conducted.
As expected, the performance of both architectures dropped drastically as the discretization became coarser (around $D=16$),  because the distance between \textit{true} embedding $\phi (\mathbf{p})$  and the approximated embedding either nearest neighbor or interpolation of \eqref{eq:trilinear} became large.   
Contrary to our expectations, one with interpolation performed slightly worse than one without interpolation.
This result may partly be attributed to the fact that the distance between the  \textit{true} embedding $\phi (\mathbf{p})$ and the approximated embedding by trilinear interpolation (LUTI) can be larger than that of the nearest-neighbor (LUT) where the linear assumption was not satisfied.

\prg{LUT-MLP E2E.}
This variant used discretized input points on $\mathbb{D}^3$ similar to LUT-MLP Approx., but it was trained E2E.
The network that used coarser lattice performed poorly, as expected, but the drop in performance was rather moderate compared with naïve approximation (LUT-MLP Approx.) because the network parameters can adapt to the coarse input. 
% The discretization loses information on input similar to the case of CNN-based architecture \cite{tran2015learning} with coarse voxelization performs poorly.     
% As expected, the performance dropped as the discretization became coarser (around $\text{D}=16$).

\prg{LUTI-MLP E2E.}
This variant is the proposed architecture, which incorporates the LUTI-MLP and trains the network E2E.
For this variant, we used a pre-trained parameter from the MLP at $100$ epochs and then trained for another $100$ epochs\footnote{The primary purpose of pre-training was to see the evolution of the embedding space in \Sec \ref{subsec:exp_vis}. In practice, this has little impact on accuracy (\Supp{\ref{suppsec:scrach}}).}.
This variant performed slightly better than the original architecture for a wide range of $D$ and outperformed other variants by a large margin.
Surprisingly, the variant works as well as those that use MLP embedding even for the lowest lattice resolution ($D$=2).
This may seem counterintuitive, but note that even for the coarsest resolution, precise input location are stored  in the point-wise embedding vector, and this information is likely to be preserved on the global feature $\mathbf{a}$ after $\max$ thanks to the irregular interpolation of \eqref{eq:ccv},
coarse lattice limits the expressivity, but not the spatial resolution or the number of active set $\mathbf{P}_{\text{a}}$.
We suspect that this restricted structure by local linearity somehow plays the role of regularizer, 
resulting in higher performance than PointNet when moderately coarse lattices are used. 

\subsection{Visualizing the PointNet Embedding Space}
\label{subsec:exp_vis}
\begin{figure}[ht]

% first image
\begin{subfigure}[t]{.47\textwidth}
\centering
\includegraphics[width=1\columnwidth]{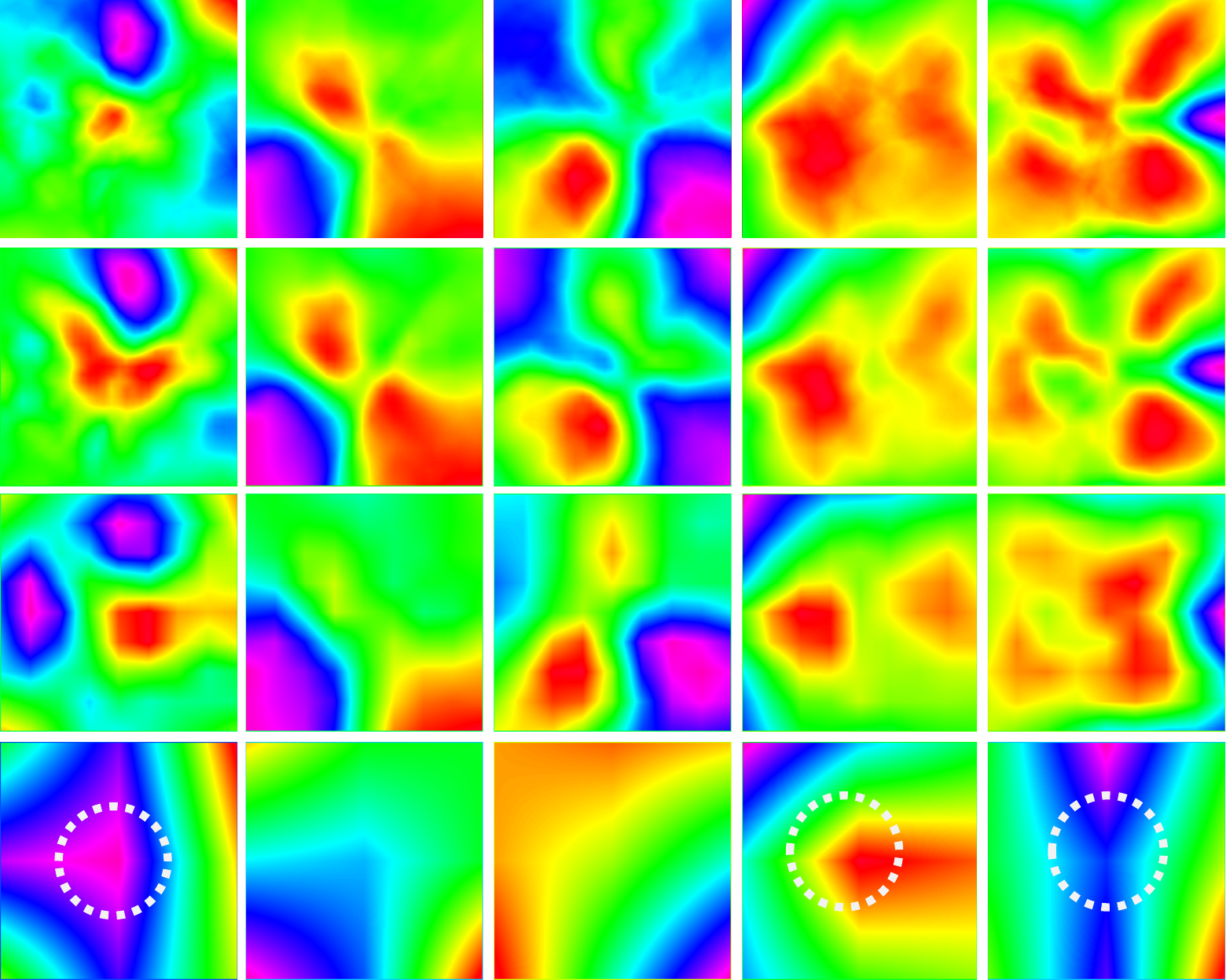}
\caption{
\textbf{LUTI$_{\text{uni}}$-MLP.} 
}
\label{fig:vis_emb_LUT_uni}
\end{subfigure}
\hspace{0.06\columnwidth}
% second image
\begin{subfigure}[t]{.47\columnwidth}
\centering
\includegraphics[width=1\columnwidth]{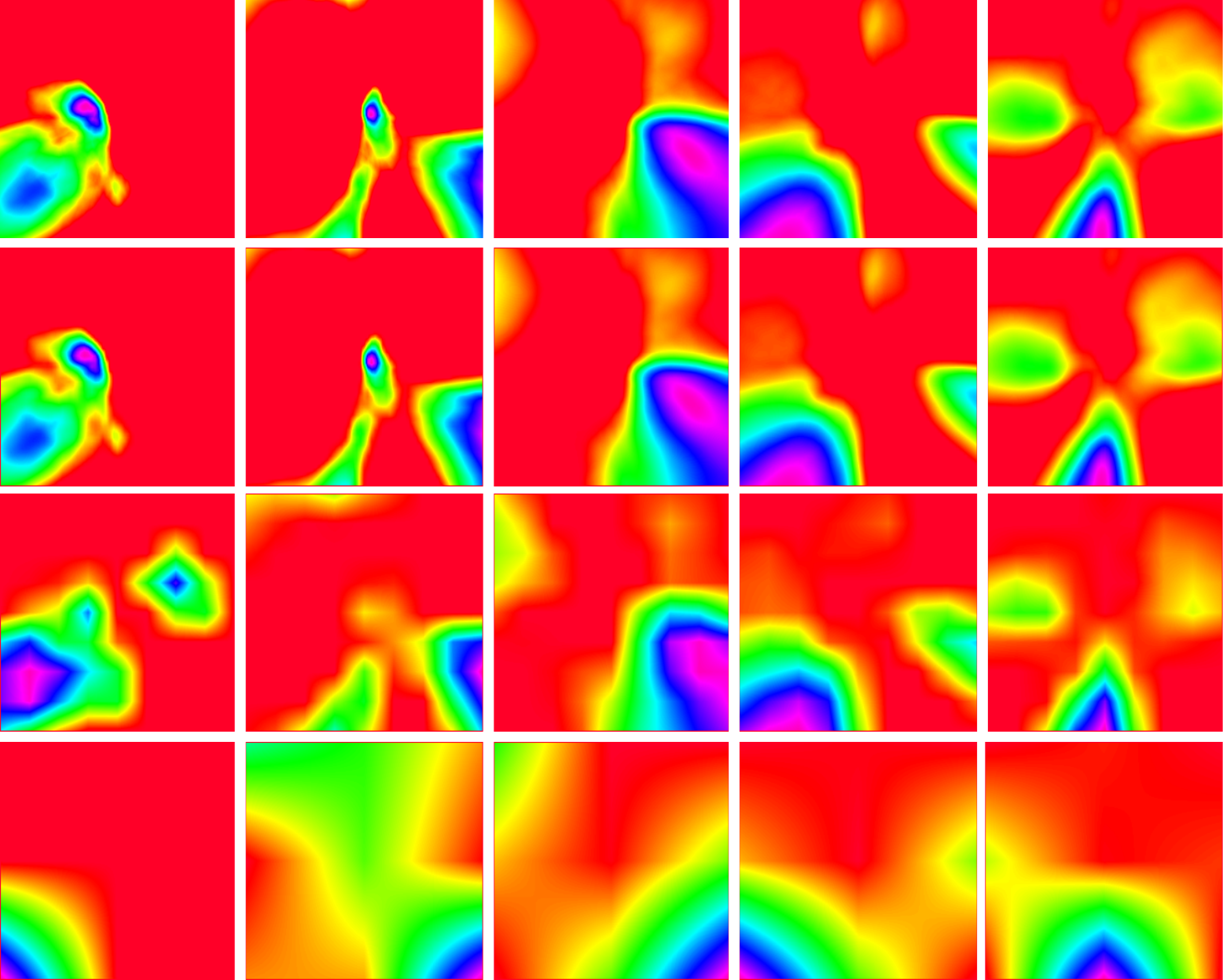}  
\caption{
\textbf{LUTI$_{\text{irr}}$-MLP.} 
}
\label{fig:vis_emb_LUT_irr}
\end{subfigure}
\caption{
\label{fig:vis_LUTI}
\textbf{Visualization of the trained embedding feature.}
From top to bottom, MLP (Equivalent to PointNet), LUTI-MLP with discretizations $D=32, 8$ and $2$.
Five randomly selected channels of the slice on the $z=0$ plane are shown.
(a): Maximum is observed only on edge (dashed circle) in case of LUTI$_{\text{uni}}$; (b): conversely, we can see irregularly arranged peaks in case of $\text{LUTI}_{\text{irr}}$. 
Each LUTI variant used the same pre-trained model from MLP for $100$ epochs using the ModelNet40 classification task.
}
\label{fig:vs_emb}
\vspace{-4mm}%Put here to reduce too much white space after your table 

\end{figure}

% We investigated the embedding space trained by LUTI-MLP at different lattice resolution.
In \Fig\ref{fig:vis_LUTI}, we visualize the trained embedding space of  LUTI\footnote{More visualizations of the embedding space are shown in \Supp{\ref{suppsec:vis_mlp},\ref{suppsec:vis_direct}}.}.
We can visually inspect that the maximum appear only on edge in case of $\text{LUTI}_{\text{uni}}$ (\Fig\ref{fig:vis_emb_LUT_uni}); conversely,  we can observe peaks scattered across at non-lattice position in case of $\text{LUTI}_{\text{irr}}$ (\Fig\ref{fig:vis_emb_LUT_irr}). 
The difference is more apparent when the lattice resolution are coarse.
As discussed in \Sec\ref{subsec:embedding}, we suspect this attribute to the good performance when coarse lattice are used. 
\section{Related Literature} 
\label{sec:related}
\prg{Neural Network Speedup.}
There is a wide range of research efforts for speeding up network inference, 
such as weight pruning \cite{he2018amc,Han2015DeepCC}, distillation \cite{hinton2015distilling,wong2016sequence}, hardware-efficient quantization \cite{wang2019haq,lin2016fixed}, and compact network designing \cite{howard2019searching,ma2018shufflenet}.
Existing approaches reduce the computational load of layers of matrix-vector products, while the proposed approach demands no such layers by construction, and instead is comprised of much more efficient LUT followed by  interpolation.
% LUTI-MLP is probably more efficient than others in many cases because there are no layers of matrix- vector products, but its special structure makes it difficult to handle high-dimensional inputs as they are [R2-1], so its applicability is not as general as others. We considered the comparison with basic MLP to be of utmost importance. We'll add a fair discussion of the above limitations to Sec.4.
% We present a new line of research toward very efficient point-feature embedding.
% for the field of point-data processing.

\prg{LUT Realization of Point Feature Embedding.}
Utilization of the LUT for point-feature embedding was first discussed in \cite{sekikawa2019eventnet} for asynchronous event-data \cite{4444573} processing.
The author of \cite{sekikawa2019eventnet} utilized the discrete nature of the event data (pixel location and polarity), and  pre-compute the input/output relationship of the embedding MLP on a LUT; this simple method cannot be used to process continuous input as in the proposed method.
LUTI-MLP is easily applicable to \cite{sekikawa2019eventnet} to drastically reduce the memory footprint for a high-resolution camera.

\prg{Point set processing for  real-world data.}
Many of the recent models for point set processing have been reported to perform well on complex real-world data by concatenating PointNet like structures into multiple layers \cite{qi2017pointnetplusplus,thomas2019kpconv,dgcnn,wang2018local,landrieu2018large}.
The evaluation of these models by adopting LUTI-MLP on real-world dataset is an important research topic. 
However, the input dimension of LUTI-MLP is limited by memory constraints (see Supp.A), so some ingenuity is needed to be used in these multi-layer models (e.g., adding layers to reduce the number of feature dimensions, such as the T-Net in \Fig S1).
Whereas, models, such as Pointpillars \cite{lang2019pointpillars} and FrustumPointNet\cite{qi2018frustum}, use a single-layer PointNet as its backbone which also works well in real-world applications.
Our main interest in this study is evaluating the distinct idea of LUTI-MLP in its most basic setup (model/data).
The application of LUTI-MLP to recent high-performance models (either multi-layer/single-layer PointNet backbone) and its evaluation on real-world datasets will be presented as a subsequent work.

\section{Conclusion and Future Work}
\label{sec:conclusion}
In this paper, we proposed a novel point-feature embedding method realized by linear combination of the basis function; that is pre-computed on a substantially irregular lattice.
With an intensive architectural analysis, we showed that the proposed LUTI-MLP speeds up the embedding by about 100$\times$ without performance degeneration.
Furthermore, the proposed formulation enables efficient analytical Jacobian computation, which has a wide range of practical applications.
% The results also infers that LUTI-MLP is effective for largely different types of network and applications.
A future research direction would be to apply this method to a large-scale real-world application or to extends the idea of LUTI-MLP into hierarchical \cite{qi2017pointnetplusplus} or graph \cite{thomas2019kpconv,dgcnn,wang2018local,landrieu2018large} structures.

\newpage
\clearpage
{\small
\bibliographystyle{splncs04}
\bibliography{bib}

\begin{thebibliography}{10}
\providecommand{\url}[1]{\texttt{#1}}
\providecommand{\urlprefix}{URL }
\providecommand{\doi}[1]{https://doi.org/#1}

\bibitem{xadams2010fast}
Adams, A., Baek, J., Davis, M.A.: Fast high-dimensional filtering using the
  permutohedral lattice. In: Computer Graphics Forum. vol.~29, pp. 753--762.
  Wiley Online Library (2010)

\bibitem{aoki2019pointnetlk}
Aoki, Y., Goforth, H., Srivatsan, R.A., Lucey, S.: Pointnetlk: Robust \&
  efficient point cloud registration using pointnet. In: Proceedings of the
  IEEE Conference on Computer Vision and Pattern Recognition. pp. 7163--7172
  (2019)

\bibitem{baker2004lucas}
Baker, S., Matthews, I.: Lucas-kanade 20 years on: A unifying framework.
  International journal of computer vision  \textbf{56}(3),  221--255 (2004)

\bibitem{shapenet2015}
Chang, A.X., Funkhouser, T., Guibas, L., Hanrahan, P., Huang, Q., Li, Z.,
  Savarese, S., Savva, M., Song, S., Su, H., Xiao, J., Yi, L., Yu, F.:
  {ShapeNet: An Information-Rich 3D Model Repository}. Tech. Rep.
  arXiv:1512.03012 [cs.GR], Stanford University --- Princeton University ---
  Toyota Technological Institute at Chicago (2015)

\bibitem{choy20163d}
Choy, C.B., Xu, D., Gwak, J., Chen, K., Savarese, S.: 3d-r2n2: A unified
  approach for single and multi-view 3d object reconstruction. In: European
  conference on computer vision. pp. 628--644. Springer (2016)

\bibitem{cciccek20163d}
{\c{C}}i{\c{c}}ek, {\"O}., Abdulkadir, A., Lienkamp, S.S., Brox, T.,
  Ronneberger, O.: 3d u-net: learning dense volumetric segmentation from sparse
  annotation. In: International conference on medical image computing and
  computer-assisted intervention. pp. 424--432. Springer (2016)

\bibitem{danzer20192d}
Danzer, A., Griebel, T., Bach, M., Dietmayer, K.: 2d car detection in radar
  data with pointnets. In: 2019 IEEE Intelligent Transportation Systems
  Conference (ITSC). pp. 61--66. IEEE (2019)

\bibitem{gross2019alignnet}
Gro{\ss}, J., O{\v{s}}ep, A., Leibe, B.: Alignnet-3d: Fast point cloud
  registration of partially observed objects. In: 2019 International Conference
  on 3D Vision (3DV). pp. 623--632. IEEE (2019)

\bibitem{Han2015DeepCC}
Han, S., Mao, H., Dally, W.J.: Deep compression: Compressing deep neural
  network with pruning, trained quantization and huffman coding. CoRR
  \textbf{abs/1510.00149} (2015)

\bibitem{he2018amc}
He, Y., Lin, J., Liu, Z., Wang, H., Li, L.J., Han, S.: Amc: Automl for model
  compression and acceleration on mobile devices. In: Proceedings of the
  European Conference on Computer Vision (ECCV). pp. 784--800 (2018)

\bibitem{hinton2015distilling}
Hinton, G., Vinyals, O., Dean, J.: Distilling the knowledge in a neural
  network. stat  \textbf{1050}, ~9 (2015)

\bibitem{howard2019searching}
Howard, A., Sandler, M., Chu, G., Chen, L.C., Chen, B., Tan, M., Wang, W., Zhu,
  Y., Pang, R., Vasudevan, V., et~al.: Searching for mobilenetv3. In:
  Proceedings of the IEEE International Conference on Computer Vision. pp.
  1314--1324 (2019)

\bibitem{ioffe2015batch}
Ioffe, S., Szegedy, C.: Batch normalization: Accelerating deep network training
  by reducing internal covariate shift. In: International Conference on Machine
  Learning. pp. 448--456 (2015)

\bibitem{jaderberg2015spatial}
Jaderberg, M., Simonyan, K., Zisserman, A., et~al.: Spatial transformer
  networks. In: Advances in neural information processing systems. pp.
  2017--2025 (2015)

\bibitem{landrieu2018large}
Landrieu, L., Simonovsky, M.: Large-scale point cloud semantic segmentation
  with superpoint graphs. In: Proceedings of the IEEE Conference on Computer
  Vision and Pattern Recognition. pp. 4558--4567 (2018)

\bibitem{lang2019pointpillars}
Lang, A.H., Vora, S., Caesar, H., Zhou, L., Yang, J., Beijbom, O.:
  Pointpillars: Fast encoders for object detection from point clouds. In:
  Proceedings of the IEEE Conference on Computer Vision and Pattern
  Recognition. pp. 12697--12705 (2019)

\bibitem{4444573}
Lichtsteiner, P., Posch, C., Delbruck, T.: A 128$\times$128 120 db 15$\mu$s
  latency asynchronous temporal contrast vision sensor. IEEE Journal of
  Solid-State Circuits  \textbf{43}(2),  566--576 (Feb 2008).
  \doi{10.1109/JSSC.2007.914337}

\bibitem{lin2016fixed}
Lin, D., Talathi, S., Annapureddy, S.: Fixed point quantization of deep
  convolutional networks. In: International Conference on Machine Learning. pp.
  2849--2858 (2016)

\bibitem{liu2019point}
Liu, Z., Tang, H., Lin, Y., Han, S.: Point-voxel cnn for efficient 3d deep
  learning. In: Advances in Neural Information Processing Systems. pp. 963--973
  (2019)

\bibitem{lu19}
Lu, W., Wan, G., Zhou, Y., Fu, X., Yuan, P., Song, S.: Deepicp: An end-to-end
  deep neural network for 3d point cloud registration. arXiv:1905.04153  (2019)

\bibitem{ma2018shufflenet}
Ma, N., Zhang, X., Zheng, H.T., Sun, J.: Shufflenet v2: Practical guidelines
  for efficient cnn architecture design. In: Proceedings of the European
  Conference on Computer Vision (ECCV). pp. 116--131 (2018)

\bibitem{maturana2015voxnet}
Maturana, D., Scherer, S.: Voxnet: A 3d convolutional neural network for
  real-time object recognition. In: 2015 IEEE/RSJ International Conference on
  Intelligent Robots and Systems (IROS). pp. 922--928. IEEE (2015)

\bibitem{paszke2017automatic}
Paszke, A., Gross, S., Chintala, S., Chanan, G., Yang, E., DeVito, Z., Lin, Z.,
  Desmaison, A., Antiga, L., Lerer, A.: Automatic differentiation in {PyTorch}.
  In: NIPS Autodiff Workshop (2017)

\bibitem{qi2019deep}
Qi, C.R., Litany, O., He, K., Guibas, L.J.: Deep hough voting for 3d object
  detection in point clouds. In: Proceedings of the IEEE International
  Conference on Computer Vision. pp. 9277--9286 (2019)

\bibitem{qi2018frustum}
Qi, C.R., Liu, W., Wu, C., Su, H., Guibas, L.J.: Frustum pointnets for 3d
  object detection from rgb-d data. In: Proceedings of the IEEE Conference on
  Computer Vision and Pattern Recognition. pp. 918--927 (2018)

\bibitem{qi2017pointnet}
Qi, C.R., Su, H., Mo, K., Guibas, L.J.: Pointnet: Deep learning on point sets
  for 3d classification and segmentation. Proc. Computer Vision and Pattern
  Recognition (CVPR), IEEE  (2017)

\bibitem{qi2017pointnetplusplus}
Qi, C.R., Yi, L., Su, H., Guibas, L.J.: Pointnet++: Deep hierarchical feature
  learning on point sets in a metric space. arXiv preprint arXiv:1706.02413
  (2017)

\bibitem{riegler2017octnet}
Riegler, G., Osman~Ulusoy, A., Geiger, A.: Octnet: Learning deep 3d
  representations at high resolutions. In: Proceedings of the IEEE Conference
  on Computer Vision and Pattern Recognition. pp. 3577--3586 (2017)

\bibitem{roynard2018classification}
Roynard, X., Deschaud, J.E., Goulette, F.: Classification of point cloud scenes
  with multiscale voxel deep network. arXiv preprint arXiv:1804.03583  (2018)

\bibitem{rusinkiewicz2001efficient}
Rusinkiewicz, S., Levoy, M.: Efficient variants of the icp algorithm. In:
  Proceedings Third International Conference on 3-D Digital Imaging and
  Modeling. pp. 145--152. IEEE (2001)

\bibitem{Sarode2019PCRNetPC}
Sarode, V., Li, X., Goforth, H., Aoki, Y., Srivatsan, R.A., Lucey, S., Choset,
  H.: Pcrnet: Point cloud registration network using pointnet encoding. ArXiv
  \textbf{abs/1908.07906} (2019)

\bibitem{sekikawa2019eventnet}
Sekikawa, Y., Hara, K., Saito, H.: Eventnet: Asynchronous recursive event
  processing. In: Proceedings of the IEEE Conference on Computer Vision and
  Pattern Recognition. pp. 3887--3896 (2019)

\bibitem{su2018splatnet}
Su, H., Jampani, V., Sun, D., Maji, S., Kalogerakis, E., Yang, M.H., Kautz, J.:
  Splatnet: Sparse lattice networks for point cloud processing. In: Proceedings
  of the IEEE Conference on Computer Vision and Pattern Recognition. pp.
  2530--2539 (2018)

\bibitem{thomas2019kpconv}
Thomas, H., Qi, C.R., Deschaud, J.E., Marcotegui, B., Goulette, F., Guibas,
  L.J.: Kpconv: Flexible and deformable convolution for point clouds. arXiv
  preprint arXiv:1904.08889  (2019)

\bibitem{tran2015learning}
Tran, D., Bourdev, L., Fergus, R., Torresani, L., Paluri, M.: Learning
  spatiotemporal features with 3d convolutional networks. In: Proceedings of
  the IEEE international conference on computer vision. pp. 4489--4497 (2015)

\bibitem{wang2018local}
Wang, C., Samari, B., Siddiqi, K.: Local spectral graph convolution for point
  set feature learning. In: Proceedings of the European Conference on Computer
  Vision (ECCV). pp. 52--66 (2018)

\bibitem{wang2019haq}
Wang, K., Liu, Z., Lin, Y., Lin, J., Han, S.: Haq: Hardware-aware automated
  quantization with mixed precision. In: Proceedings of the IEEE conference on
  computer vision and pattern recognition. pp. 8612--8620 (2019)

\bibitem{Wang2019}
Wang, Q., Zhang, Y., Yuan, J., Lu, Y.: {Space-time event clouds for gesture
  recognition: From RGB cameras to event cameras}. Proceedings - 2019 IEEE
  Winter Conference on Applications of Computer Vision, WACV 2019 pp.
  1826--1835 (2019). \doi{10.1109/WACV.2019.00199}

\bibitem{wang2019deep}
Wang, Y., Solomon, J.M.: Deep closest point: Learning representations for point
  cloud registration. arXiv preprint arXiv:1905.03304  (2019)

\bibitem{Wang2019PRNet}
Wang, Y., Solomon, J.M.: Prnet: Self-supervised learning for partial-to-partial
  registration. In: 33rd Conference on Neural Information Processing Systems
  (2019)

\bibitem{dgcnn}
Wang, Y., Sun, Y., Liu, Z., Sarma, S.E., Bronstein, M.M., Solomon, J.M.:
  Dynamic graph cnn for learning on point clouds. ACM Transactions on Graphics
  (TOG)  (2019)

\bibitem{wong2016sequence}
Wong, J.H., Gales, M.: Sequence student-teacher training of deep neural
  networks  (2016)

\bibitem{wu20153d}
Wu, Z., Song, S., Khosla, A., Yu, F., Zhang, L., Tang, X., Xiao, J.: 3d
  shapenets: A deep representation for volumetric shapes. In: Proceedings of
  the IEEE conference on computer vision and pattern recognition. pp.
  1912--1920 (2015)

\bibitem{zhou2018voxelnet}
Zhou, Y., Tuzel, O.: Voxelnet: End-to-end learning for point cloud based 3d
  object detection. In: Proceedings of the IEEE Conference on Computer Vision
  and Pattern Recognition. pp. 4490--4499 (2018)

\end{thebibliography}
}
%%%%%%%%%%%%%%%% SUPPLEMENTAR
% http://laussy.org/wiki/Blog:Hacks/Supplementary_TeX_material_on_arXiv
\newpage
\clearpage

\renewcommand\thesection{\Alph{section}}
\renewcommand\thesubsection{\thesection.\Alph{subsection}}
% \pagebreak

\onecolumn
\begin{center}
  \textbf{\large Irregularly Tabulated MLP\\ for Fast Point Feature Embedding}\\
  [0.2cm]
  \textbf{Supplementary Material}\\
  [0.5cm]
  {Anonymous ECCV submission}\\
  [0.5cm]
  {Paper ID 2487}\\
  [1cm]
\end{center}

% \begin{center}
%   \textbf{\Large Tabulated MLP for Fast Point Feature Embedding}\\
%   [0.2cm]
%   \textbf{\Large Supplementary Material}\\
%   [0.5cm]
%   {\large Yusuke Sekikawa and Teppei Suzuki}\\
%   [0.5cm]
%   {\large DENSO IT Laboratory}\\
%   [1cm]
% \end{center}

\setcounter{section}{0}
\setcounter{equation}{0}
\setcounter{figure}{0}
\setcounter{table}{0}
\setcounter{page}{1}
\renewcommand{\theequation}{S\arabic{equation}}
\renewcommand{\thefigure}{S\arabic{figure}}
\renewcommand{\thetable}{S\arabic{table}}

\section{Memory Footprint Consideration}
\label{suppsec:memory}
\Tab\ref{table:memory}  compares the memory footprint of proposed LookUp Table Interpolation multi-layer perceptron (LUTI-MLP) with $\mathbb{D}^M$ lattice for a different lattice resolution $D$ and different lattice dimension $M$.
When $D=4$ and $M=3$, the size of the table is about 250 KB, which is even smaller ($2.3\times$) than that of the embedding multi-layer perceptron (MLP) used in the original PointNet \cite{qi2017pointnet} (\Fig\ref{fig:overview}, top). 
In the main paper, we focused our discussion on 3D data $\in \mathbb{R}^3$, such as a 3D geometric point cloud; thus, a $\mathbb{D}^3$ lattice whose memory requirement is $\propto D^3$.
The LUTI-MLP can be extensible to the input  $\in \mathbb{R}^M$ by using a $\mathbb{D}^M$ lattice as much as the memory for storing the basis function of \eqref{eq:lutimlp_basic} permits.
When we want to incorporate additional features, such as intensity, color, normal, and so on, the lattice dimension $M$ needs to be extended accordingly, which increases the memory footprint exponentially, that is $\propto D^M$.
For example, when we incorporate the intensity information from LiDAR data, the input point feature would be $\in \mathbb{R}^4$, and the lattice becomes $\mathbb{D}^4$.
In this case, about $16$ MB of memory is required for $D=8$ to store the basis function.
When the normal information is further incorporated, the input point feature would be $\in \mathbb{R}^6$, and the memory footprint would be about $1$ GB for $D=8$.
For such higher-dimensional inputs, it may be possible to keep the look-up table (LUT) size reasonable by utilizing a non-Euclidean lattice, such as a permutohedral lattice \cite{su2018splatnet,xadams2010fast}.
\begin{table}[h]
\caption{\label{table:memory}
\textbf{Comparison of the memory footprint of MLP and LUTI-MLP.}
We compare the memory footprint (in MB) of the LUTI-MLP with $\mathbb{D}^M$ lattice for different lattice resolution $D$ and different lattice dimension $M$.
A $4$ byte floating point is assumed for storing each parameter.
Note that the memory footprint of LUTI using uniform grid (LUTI$_{\text{uni}}$) is the same as LUTI using irregular grid (LUTI$_{\text{irr}}$)
}

\begin{center}
\begin{adjustbox}{width=1\columnwidth}
\begin{tabular}{c|c:cccccccc}
& MLP & \multicolumn{5}{c}{ LUTI with $D$ } & & & \\ 
\hline
$M$ & & 1024 & 64 & 32 & 16 & 8 & 5 & 4 & 2 \\ \hdashline
3 & 5.68E-01 & 4.19E+06 & 1.02E+03 & 1.28E+02 & 1.60E+01 & 2.00E+00 & 4.88E-01 & 2.50E-01 & 3.13E-02 \\
4 & 5.68E-01 & 4.29E+09 & 6.55E+04 & 4.10E+03 & 2.56E+02 & 1.60E+01 & 2.44E+00 & 1.00E+00 & 6.25E-02 \\
5 & 5.69E-01 & 4.40E+12 & 4.19E+06 & 1.31E+05 & 4.10E+03 & 1.28E+02 & 1.22E+01 & 4.00E+00 & 1.25E-01 \\
6 & 5.69E-01 & 4.50E+15 & 2.68E+08 & 4.19E+06 & 6.55E+04 & 1.02E+03 & 6.10E+01 & 1.60E+01 & 2.50E-01 \\
\end{tabular}
\end{adjustbox}

\end{center}
\end{table}

%%%%%%%%%%%%%%%%%%%%%%%%%%%%%%%%%%%%%%%%%%%%%%%
\newpage
\clearpage
\section{LUTI-MLP for Transformation Network}
\label{suppsec:tnet}
To get some kind of invariance against a geometric transformation of input, PointNet \cite{qi2017pointnet} also proposed the input/feature transformation mechanism called T-Net, inspired by spatial transformer network (STN) \cite{jaderberg2015spatial}.
T-Net is intended to canonicalize the input point set through an affine transformation.
In this supplementary section, we describe the detailed architecture of T-Net using LUTI-MLP for acceleration.
T-Net is actually a small version of PointNet, which is used to estimate the affine transformation matrix; this transformation is directly applied to the input points coordinates and/or intermediate features. 
In \cite{qi2017pointnet},  T-Net is used at two different layers, one for the input layer as the input transform (IT) and the other for the middle of the embedding MLP as the feature transform (FT).
T-Net (FT) needs additional care with training, because it requires orthogonal regularization for the affine parameter to get better performance; otherwise, it has been reported (\cite{qi2017pointnet}) to deteriorate the performance instead.
The need for regularization introduces an additional hyperparameter to balance the primary loss and the regularization loss.

In our preliminary experiments using vanilla PointNet on ModelNet40 \cite{wu20153d}, T-Net (IT) slightly improved the accuracy (w/o IT: 86.23, w/ IT: 87.43\%); however, we did not observe any improvement by using T-Net (IT) and T-Net (FT) (w/ IT+FT 87.24\%) even though we applied the orthogonality regularization.
Although we saw no performance improvement by using T-Net (FT), we suspect that it depends on the dataset, training procedure, or initialization of parameters. 
We think there exist situations where the canonicalization using T-Net (FT) would improve performance.
Therefore, in the following subsection, we will describe the architectures using LUTI-MLP for both types of T-Net. 

\subsection*{LUTI Accelerated T-Net}
\begin{figure}[!h]
\includegraphics[width=1\columnwidth]{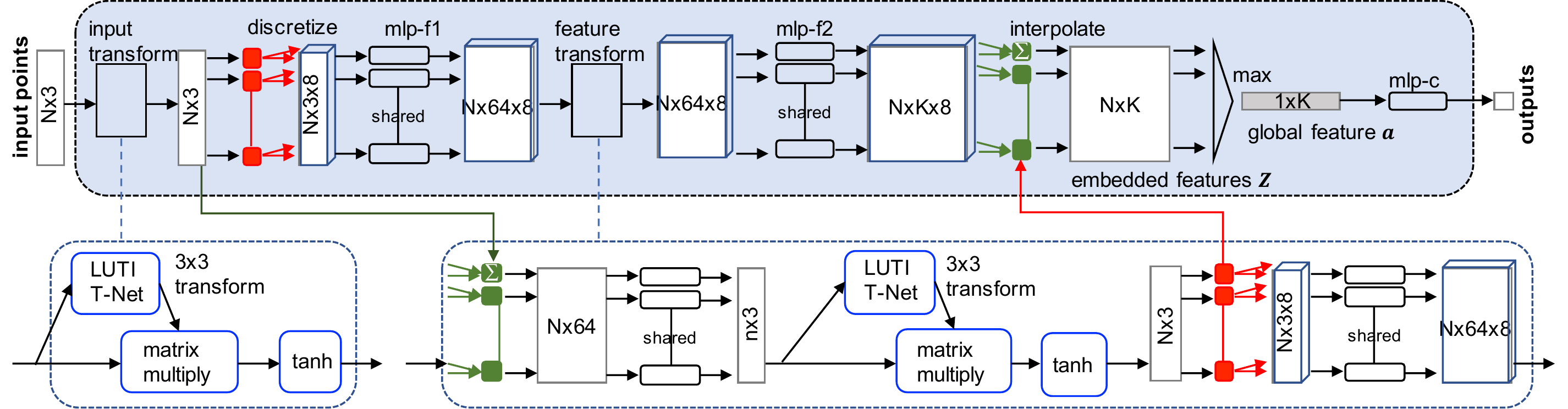}
\caption{
\textbf{Object classification network using LUTI-MLP incorporating T-Net (at training).}
To incorporate T-Net,  we modified the original T-Net (IT) by adding a $\tanh$ function to bound the output transformed coordinates.
The T-Net (FT) further includes two linear layers coupled with an interpolation/discretization module that decrease and increase the input feature to reduce the look-up table's dimensions into a manageable lattice size  ($\mathbb{D}^3$)
\label{fig:tnet}
}
\end{figure}

LUTI-MLP can accelerate T-Net with a slight modification.
The embedding MLP of T-Net can be replaced by the LUTI-MLP to speed up the computation of the transformation matrix.
\Fig\ref{fig:tnet} shows a detailed network architecture with T-Net (IT) and T-Net (FT) using LUTI-MLP.
At test time, all the embedding MLPs of the primary network, T-Net (IT), and T-Net (FT) are realized as interpolation of basis function \eqref{eq:lutimlp_basic} that is much faster than evaluating MLPs.

% PointNet with LUTI-MLP incorporating T-Net is shown.
\subsection*{T-Net for IT}
To limit the range of data feed to the LUTI-MLP on the following layer, we modified the original T-Net (IT) slightly by inserting $\tanh$ at the end of T-Net.
\Figure\ref{fig:effect_D_supp} on \Supp{\ref{suppsec:detail_analisys}} shows the additional results that follow on  \Fig\ref{fig:effect_D} from the main paper, which includes the results using T-Net (IT).
Similar to the results without T-Net (FT) described in  the main paper, the networks with T-Net (IT) accelerated by LUTI-MLP  achieved comparable performance to its counterpart that used MLP embedding.

\subsection*{T-Net for FT}
The original T-Net (FT) consumes $N\times64$ vectors to estimate  $64\times 64$ affine matrix.
But it is hard to precompute and store a 64-dimensional lattice ($\mathbb{D}^{64}$) because of the memory footprint (\Tab\ref{table:memory}), so we modified T-Net (FT) so that the dimension of the lattice became reasonably small ($\mathbb{D}^{3}$), as shown in \Fig\ref{fig:tnet}.
Our modified T-Net (FT) includes two additional linear layers that decrease and increase the input feature. 
When coupled with the interpolation/discretization layer, it makes the LUT's dimensions a manageable size ($\mathbb{D}^3$).
For the modified T-Net (FT), we did not use regularization, which encourages the matrix to be orthogonal,  because it has fewer parameters (the same as T-Net (IT)) than the original $64\times64$ affine matrix.

The object classification accuracy of LUTI$_{\text{uni}}$-MLP and  LUTI$_{\text{irr}}$-MLP incorporating both T-Net (IT) and T-Net (FT) were 87.01\% and 87.57\% respectively when $D=4$.
In our detailed architecture analysis on \Supp{\ref{suppsec:detail_analisys}}, we skipped the evaluation by varying the lattice resolution $D$ for this variant, because in the preliminary experiment using vanilla PointNet, the use of IT+FT did not show improvement as discussed earlier.
% \input{table/modelnet_ft.tex}

%%%%%%%%%%%%%%%%%%%%%%%%%%%%%%%%%%%%%%%%%%%%%%%
\newpage
\clearpage
\section{Details on the Point-Wise Classification Network}
\label{suppsec:seg}
\begin{figure}[!h]
\includegraphics[width=1\columnwidth]{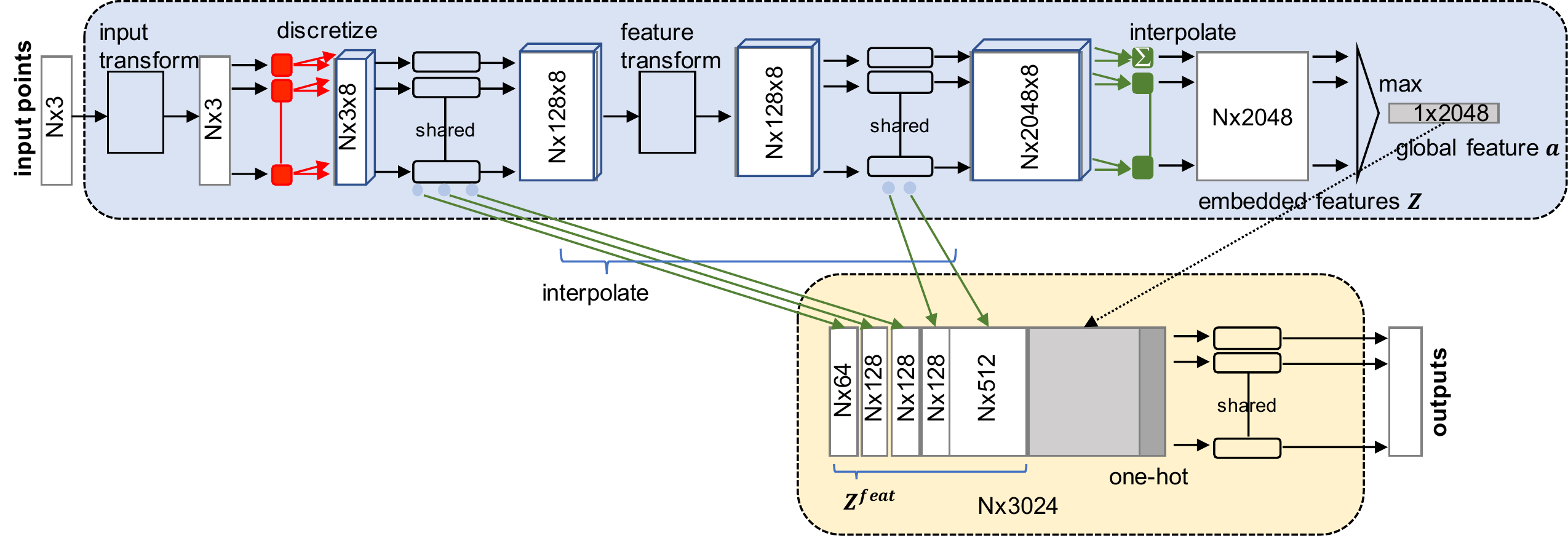}
\caption{
\textbf{Point-wise classification network using LUTI-MLP (at training).}
The network architecture is basically the same as the point-wise classification network of PointNet \cite{qi2017pointnet} for processing ShapeNet \cite{shapenet2015}, except that the embedded MLP is replaced with our LUTI-MLP.
At test time, both $\mathbf{Z}$ and $\mathbf{Z}^{feat}$ are computed as a trilinear interpolation of basis functions stored on an LUT sized $D^3\times (64+128+128+512+2048)$.
T-Net (IT) and T-Net (FT) are omitted for clarity
\label{fig:network_seg}
}
\end{figure}

The detailed network architecture used for the parts segmentation task discussed in \Sec\ref{subsec:exp_app} of the main paper are shown in \Fig\ref{fig:network_seg}.
The network is identical to the point-wise classification network used for the parts segmentation task in PointNet \cite{qi2017pointnet}, except for the LUTI structure.
The intermediate features evaluated on the discrete lattice at different layers are interpolated and then concatenated to get the point-wise embedding feature $\mathbf{Z}_{feat}$.

% \Tab \ref{table:shapenet_seg} shows additional results of parts segmentation task using ShapeNet \cite{shapenet2015}, evaluated with different discretization  $\text{D}$ that was omitted from the main paper (\Sec\ref{subsec:exp_app}, 3D Object Part Segmentation, \Tab\ref{table:shapenet_digest}). 
% The performance is equally good across different discretizations.
% These results infer that our LUTI-MLP, even with small size lattice, is also effective for point-wise estimation network (), which is largely different from the classification network (\Fig\ref{fig:overview}, bottom).
% \input{table/shapenet_seg.tex}

%%%%%%%%%%%%%%%%%%%%%%%%%%%%%%%%%%%%%%%%%%%%%%%
\newpage
\clearpage
\section{Point Set Registration Using LUTI-MLP Jacobian}
\label{suppsec:pnlk}
\begin{figure}[!h]
\includegraphics[width=1\columnwidth,trim={0 0 0 0cm},clip]{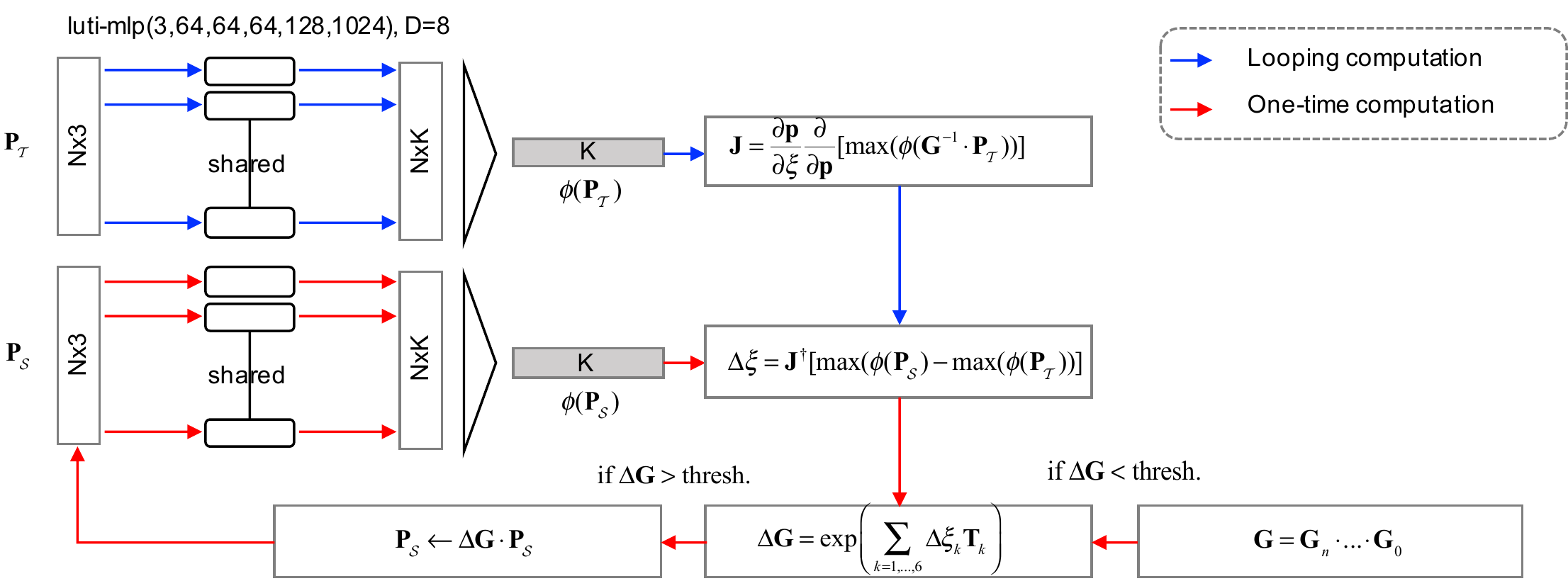}
\caption{
\textbf{Network architecture of PointNetLK using analytical Jacobian of LUTI-MLP (at test time).}
Given the source ($\mathbf{P}_{\mathcal{S}}$) and target  ($\mathbf{P}_{\mathcal{T}}$) 3D point clouds, the network incrementally estimates the 3D geometric transformation in $SE(3)$ between the sets.
The analytical Jacobian $\mathbf{J}$ is computed once using $\phi(\mathbf{P}_{\mathcal{T}})$. 
The network architectures are basically  the same as the original PointNetLK \cite{aoki2019pointnetlk}, except that the embedded MLP is replaced with our LUTI-MLP and Jacobian computation is replaced with a canonical Jacobian using the LUTI-MLP.
\label{fig:network_PNLK}
}
\end{figure}

\begin{figure}[!h]
\includegraphics[width=0.5\columnwidth]{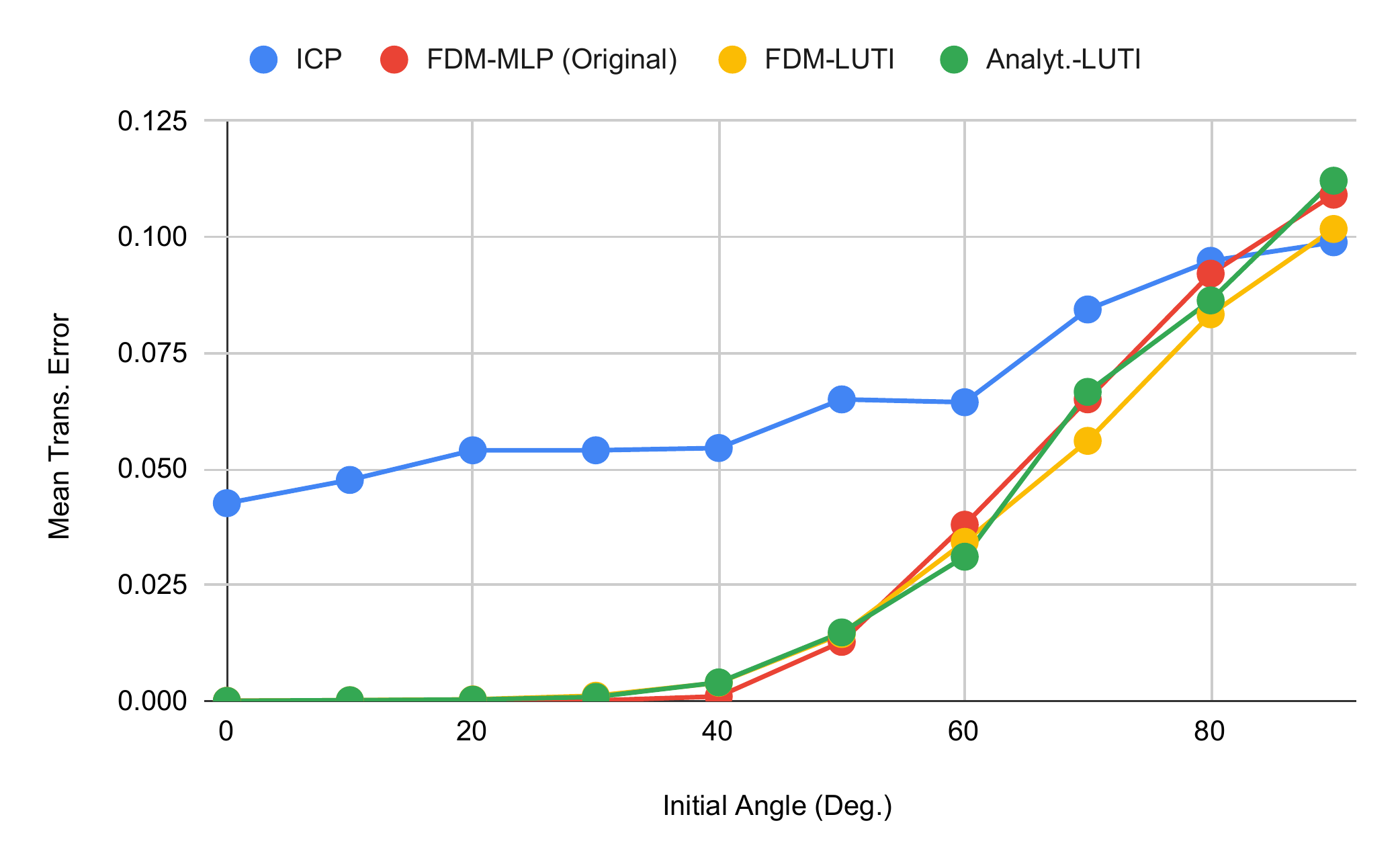}
\includegraphics[width=0.5\columnwidth]{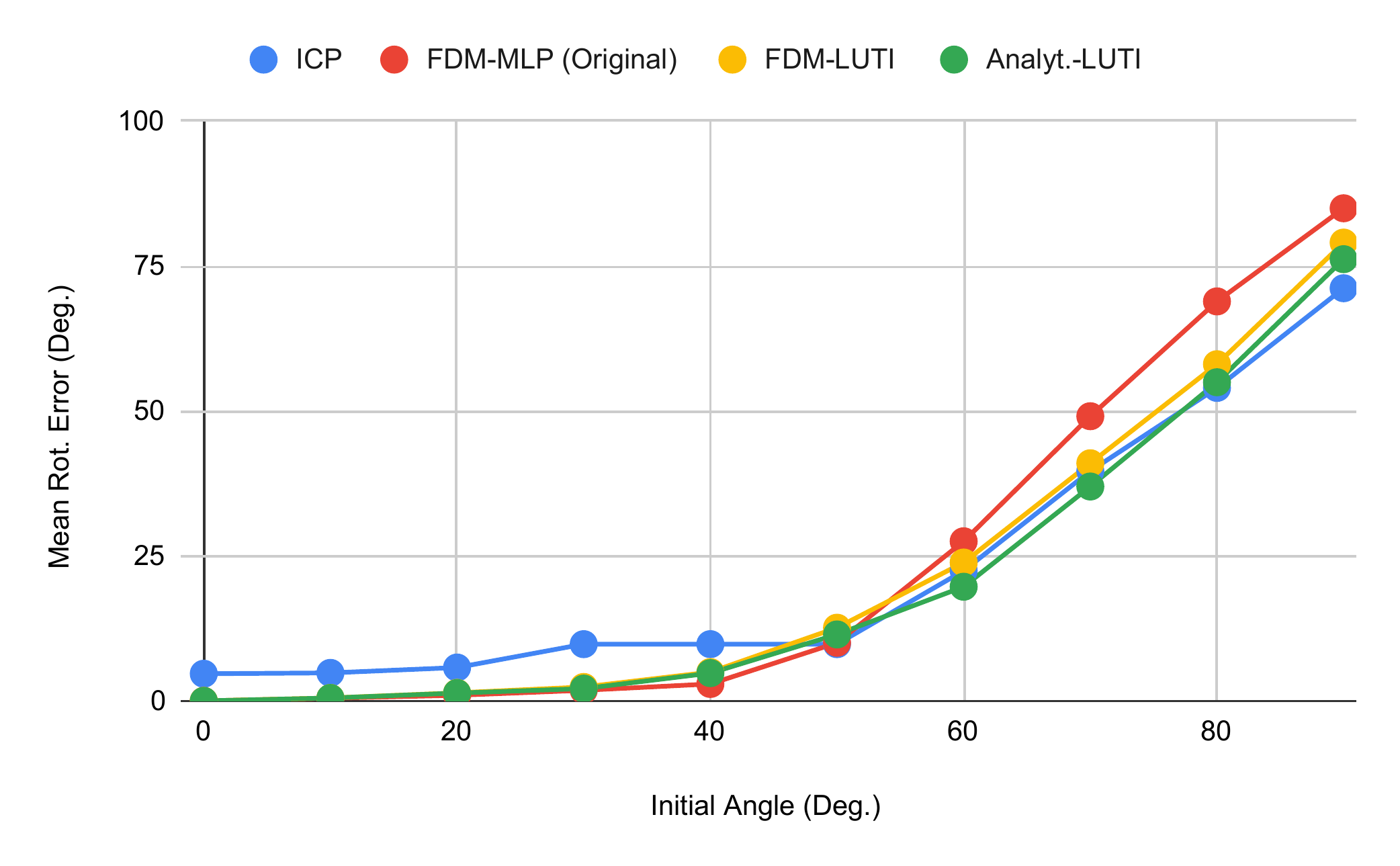}
\caption{
\label{fig:PTLK}
\textbf{Results of point cloud registration using PointNetLK accelerated by LUTI-MLP.}
The results compare the performance of the original PointNetLK (FDM-MLP), the variant using LUTI for approximate Jacobian (FDM-LUTI), and the variant using LUTI for canonical Jacobian (Analyt.-LUTI).
The results are reported for ten iterations of all architectures.
The performance is evaluated on categories unseen during training.
The LUTI-integrated networks achieved comparable performances, while also realizing an approximately $37\times$ faster overall registration process.
For this experiment using LUTI, the irregular lattice version (LUTI$_{\text{irr}}$) was used.
}
\end{figure}

\Figure\ref{fig:network_PNLK} shows the network architecture of PointNetLK at test time using the analytical Jacobian of proposed LUTI$_{\text{irr}}$-MLP.
\Figure\ref{fig:PTLK} shows the accuracy of the point cloud registration task using ModelNet40 \cite{wu20153d}, discussed in experiments in the main paper (\Sec\ref{subsec:exp_app}, 3D Point Cloud Registration).
As shown in \Fig\ref{fig:PTLK}, the registration accuracy of the three variants, FDM of MLP \cite{aoki2019pointnetlk} (FDM-MLP), FDM of LUTI$_{\text{irr}}$-MLP  (FDM-LUTI),  and analytical  Jacobian of LUTI$_{\text{irr}}$-MLP  (Analyt.-LUTI) are almost the same across all ranges of initial pose alignments (translation/rotation).
For this experiment, we followed the protocol of PointNetLK \cite{aoki2019pointnetlk} by using their published source code\footnote{https://github.com/hmgoforth/PointNetLK} with modified feature embedding part and the Jacobian computation part using the LUTI-MLP (\Fig\ref{fig:network_PNLK}).
We did not evaluate the registration accuracy of the variant using the analytical Jacobian of MLP (Analyt.-MLP), because the training and testing of this variant are computationally demanding (see \Tab\ref{table:speed_jac}).

%%%%%%%%%%%%%%%%%%%%%%%%%%%%%%%%%%%%%%%%%%%%%%%
\newpage
\clearpage
\section{Direct LUT/LUTI training}
\label{suppsec:direct}
The proposed LUTI-MLP architecture couples an MLP and an LUT in a specific manner.
In principle, it is possible to train the parameter  $\mathbf{W}$ on LUT directly, without an MLP, by using errors from a classification network.
It can be done either directly (LUT-Direct) or directly through the interpolation of \eqref{eq:trilinear} (LUTI-Direct).
These variants do not incorporate MLP as a proxy for training the parameter $\mathbf{W}$ on the LUT, but instead, directly trains $\mathbf{W}$ using error signals from upper layers.

\subsection*{LUT-Direct}
In the case of LUT-Direct, an element of a table (network parameter $\mathbf{W}$) correspond to an input point coordinate is directly updated using the backpropagated classification error.

\subsection*{LUTI-Direct} 
LUTI-Direct is equivalent to training the LUTI-MLP network at test time (\Fig \ref{fig:overview}, middle) that does not use MLP.
The elements of a table (network parameter $\mathbf{W}$) correspond to an input coordinate, and its neighbors (8 neighbors in the case of trilinear interpolation) are directly updated using the backpropagated classification error.

\subsection*{Regularization} 
In these architectures, elements of the table $\mathbf{W}$, which had no corresponding input, receive no gradient signal for training; thus, it is difficult to train when the spatial resolution of the table is fine because the gradients are sparse and only small portions of $\mathbf{W}$ are updated. 
The sparse gradient signal may negatively affect the generalization to unseen point coordinates.
To help update the network parameter $\mathbf{W}$ where inputs are not available, we regularized $\mathbf{W}$ with a total variation (TV).
We used regularization using the TV  with $p$ norm (TVL1 for $p=1$, TVL2 for $p=2$) for parameter  $\mathbf{W}$ on the table. 
The TV was evaluated using  3D spatial neighbor $\mathcal{N}$  as
\begin{equation}
T V(\mathbf{W})=\sum_{i, k \in \mathcal{N}}\left\|\mathbf{w}_{i}-\mathbf{w}_{j}\right\|^{p}.
\end{equation}
It was used in combination with standard classification objective $l_{CE}$.
Then, the loss function becomes
\begin{equation}
l=l_{CE} + \lambda T V(\mathbf{W}),
\end{equation}
where we used $\lambda=1.0$ for experiments in  \Supp{\ref{suppsec:detail_analisys}}.
The results on the ModelNet40 classification task are summarized in \Fig\ref{fig:effect_D_supp} (LUT-Direct, LUTI-Direct).
Trained embedding spaces are displayed in \Fig\ref{fig:vis_emb_direct_supp}.

%%%%%%%%%%%%%%%%%%%%%%%%%%%%%%%%%%%%%%%%%%%%%%%
\newpage
\clearpage
\section{Detailed Results from Architecture Design Analysis}
\label{suppsec:detail_analisys}
\begin{figure}[!ht]
\caption{
\label{fig:effect_D_supp}
\textbf{Detailed architecture design analysis.}
Classification accuracy from the object classification task on the Model-Net40.
Lines without markers represent the baseline results using MLP as embedding.
Our network with end-to-end (E2E) trained LUTI-MLP embedding showed comparable or slightly better results against a network using MLP embedding (PointNet) across all ranges of $D$, while achieving a significant speedup.
See \Sec\ref{sec:exp}, \Supp{\ref{suppsec:tnet}}, and \Supp{\ref{suppsec:direct}} for explanations of each algorithm.
The metric is overall classification accuracy
}
\includegraphics[width=1.0\columnwidth]{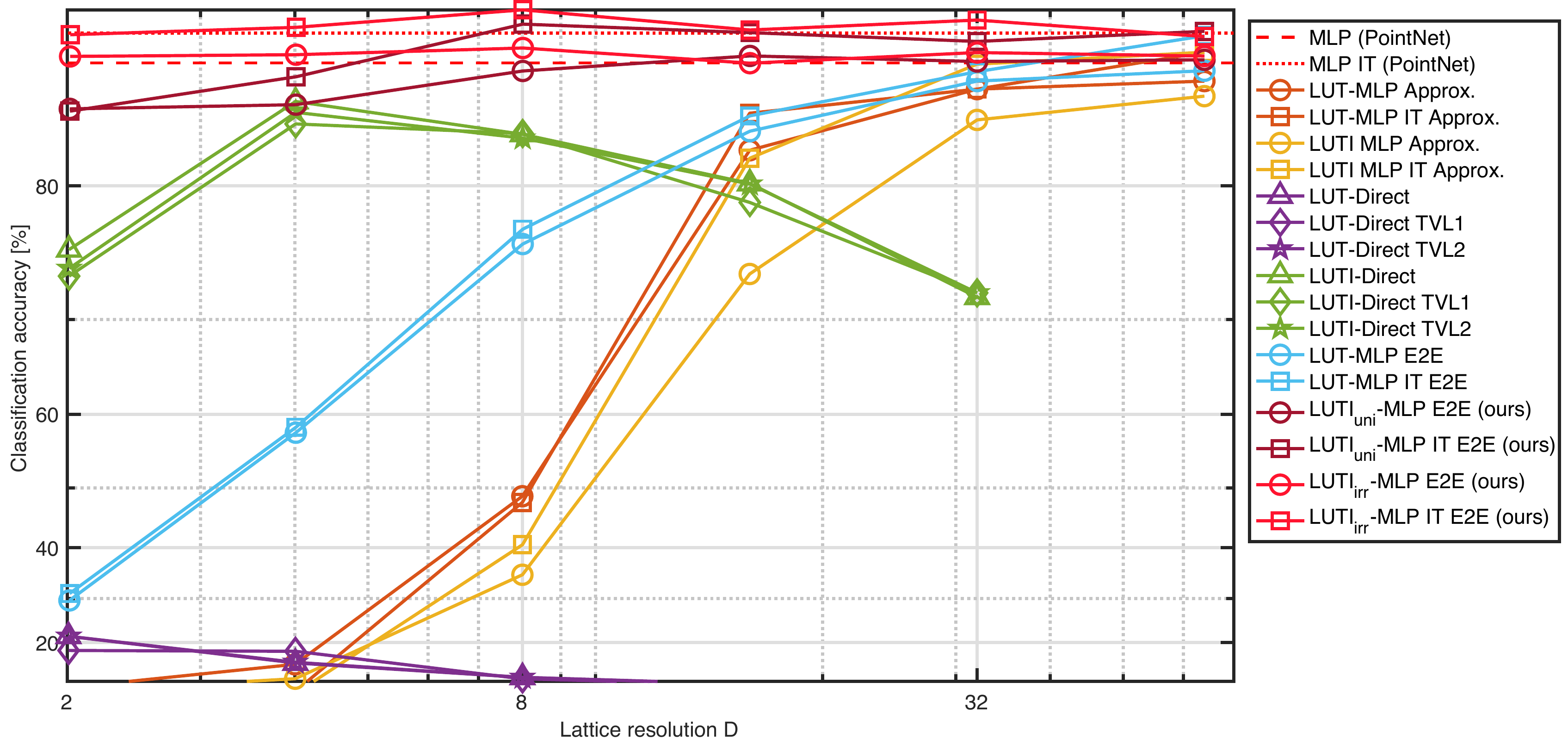}
\end{figure}

In this supplemental section, we report additional results using T-Net (\Supp{\ref{suppsec:tnet}}), and LUT(I)-Direct (\Supp{\ref{suppsec:direct}}).
\Figure\ref{fig:effect_D_supp} summarizes the results.
MLP IT (PointNet) is a re-implementation of PointNet with a slight modification on T-Net (IT), discussed in \Supp{\ref{suppsec:tnet}}, which we consider to be the baseline architecture for variants using T-Net (IT).
For all architectures except LUT-Direct and LUTI-Direct, we experimented using variants with T-Net (IT).

% \begin{figure}[!h]

\subsection*{Comparison Using T-Net (IT)}
In summary, we got similar results in the case where T-Net (IT) was incorporate. Variants using LUTI$_{\text{uni}}$-MLP/LUTI$_{\text{irr}}$-MLP embedding performed equally well as those using MLP embedding when the fine lattice was used; variants using LUTI$_{\text{irr}}$-MLP embedding performed slightly better than MLP embedding when the coarse lattice was used.

\subsection*{LUT-Direct/LUTI-Direct\footnote{These two variants were not evaluated for \textit{D} = 64 because of the limitations of our GPU memory.}}
Both LUT-Direct and LUTI-Direct perform poorly at fine lattice resolutions, even with TV regularization.
We suspect this is because even if the parameters on the LUT receive the gradients with regard to  the regularization, the gradients from the regularizer do not directly improve classification accuracy.
Surprisingly, however, LUTI-Direct performed reasonably well with a coarse lattice resolution.
This might be because the error signals are effectively propagated to a large portion of parameter $\mathbf{W}$ when the lattice resolution is coarse, thanks to the interpolation of LUTI.
In other words, the interpolation of the LUTI helps the error signal flow into the neighboring element with an appropriate weight according to the proximity of the input to the neighbor lattice, and the coarse table structure itself works as a regularizer.
Although LUTI-Direct with a specific lattice resolution performed better than expected, its best performance
was still 1–2\% worse than the proposed LUTI$_{\text{uni}}$-MLP that had the same network structure at test time.
This fact suggests that the use of MLP coupled with an table itself provides a suitable method for training the LUT for point-feature embedding using interpolation.

% It may because the gradient signal become sparse, and it is difficult to train when the spatial resolution of the table is fine.
% The comparison between LUTI$_{\text{uni}}$-MLP E2E and LUTI-Direct (\Supp{\ref{suppsec:direct}}), which has the same structure at test time, suggests that the use of MLP coupled with LUT itself provides a suitable training method for training LUT for interpolation of point-feature embedding.

%%%%%%%%%%%%%%%%%%%%%%%%%%%%%%%%%%%%%%%%%%%%%%%
\newpage
\clearpage
\section{Training PointNet with LUTI-MLP from Scratch}
\label{suppsec:scrach}
The results of end-to-end (E2E) trained PointNet with LUTI-MLP (LUTI$_{\text{uni}}$-MLP and LUTI$_{\text{irr}}$-MLP in \Fig\ref{fig:effect_D}) in the main paper used a pre-trained model (at $100$ epochs) from vanilla PointNet (using MLP embedding) for initializing the network parameters.
The primary purpose of the pre-training was to see the evolution of the embedding feature at different lattice resolution (\Sec \ref{subsec:exp_vis}). 
\Tab\ref{tab:scrach_vs_pretrain} compares the object classification accuracy on ModelNet40 \cite{wu20153d} from the PointNet with LUTI$_{\text{irr}}$-MLP trained from scratch and ones using the pre-training.
For all ranges of lattice resolution $D$ in our experiments, the performance of network trained from scratch was comparable to those trained using the pre-trained network.
As this table implies, the pre-training has little impact on classification accuracy, at least for this application.
\begin{table}[h]
\caption{
\label{tab:scrach_vs_pretrain}
\textbf{Comparison of object classification accuracy of PointNet using LUTI$_{\text{irr}}$-MLP  with and without pre-training. }
The classification accuracy of the network without pre-training is comparable to pre-trained network.
The results from different lattice resolutions $D$ are shown
}
\begin{center}
\begin{tabular}{ccc|cccccc}
\multicolumn{3}{c}{ Algorithm } & \multicolumn{5}{c}{ D } \\
\hdashline
& & IT & 64 & 32 & 16 & 8 & 4 & 2\\ 
\hline
\multirow{2}{*}{ w/ pre-train } 
& LUTI$_{\text{irr}}$- MLP E2E &   
& 86.51\% & 86.66\% & 86.22\% & 86.84\% & 86.57\% & 86.50\% \\
& LUTI$_{\text{irr}}$-MLP E2E & \checkmark                     
& 87.29\% & 87.91\% & 87.54\% & 88.28\% & 87.64\% & 87.36\% \\
\\ \hdashline
\multirow{2}{*}{ w/o pre-train } 
& LUTI$_{\text{irr}}$- MLP E2E &            
& 86.24\% & 86.56\% & 86.53\% & 86.82\% & 86.21\% & 86.36\% \\
& LUTI$_{\text{irr}}$-MLP E2E & \checkmark  
& 87.19\% & 87.99\% & 87.54\% & 88.28\% & 88.14\% & 87.79\% \\
\end{tabular}

\end{center}
\end{table}

%%%%%%%%%%%%%%%%%%%%%%%%%%%%%%%%%%%%%%%%%%%%%%%
\newpage
\clearpage
\section{Visualization of Embedding Space of LUTI-MLP}
\label{suppsec:vis_mlp}
\Figure\ref{fig:vis_emb_mlp_supp} shows more results of the trained LUTI$_{\text{uni}}$-MLP and LUTI$_{\text{irr}}$-MLP embedding feature space (\Fig \ref{fig:vis_LUTI} in the main paper) in for more diverse lattice resolutions $D$ and channels.
\begin{figure}[ht]

% first image
\begin{subfigure}[t]{1\textwidth}
\centering
\includegraphics[width=1\columnwidth]{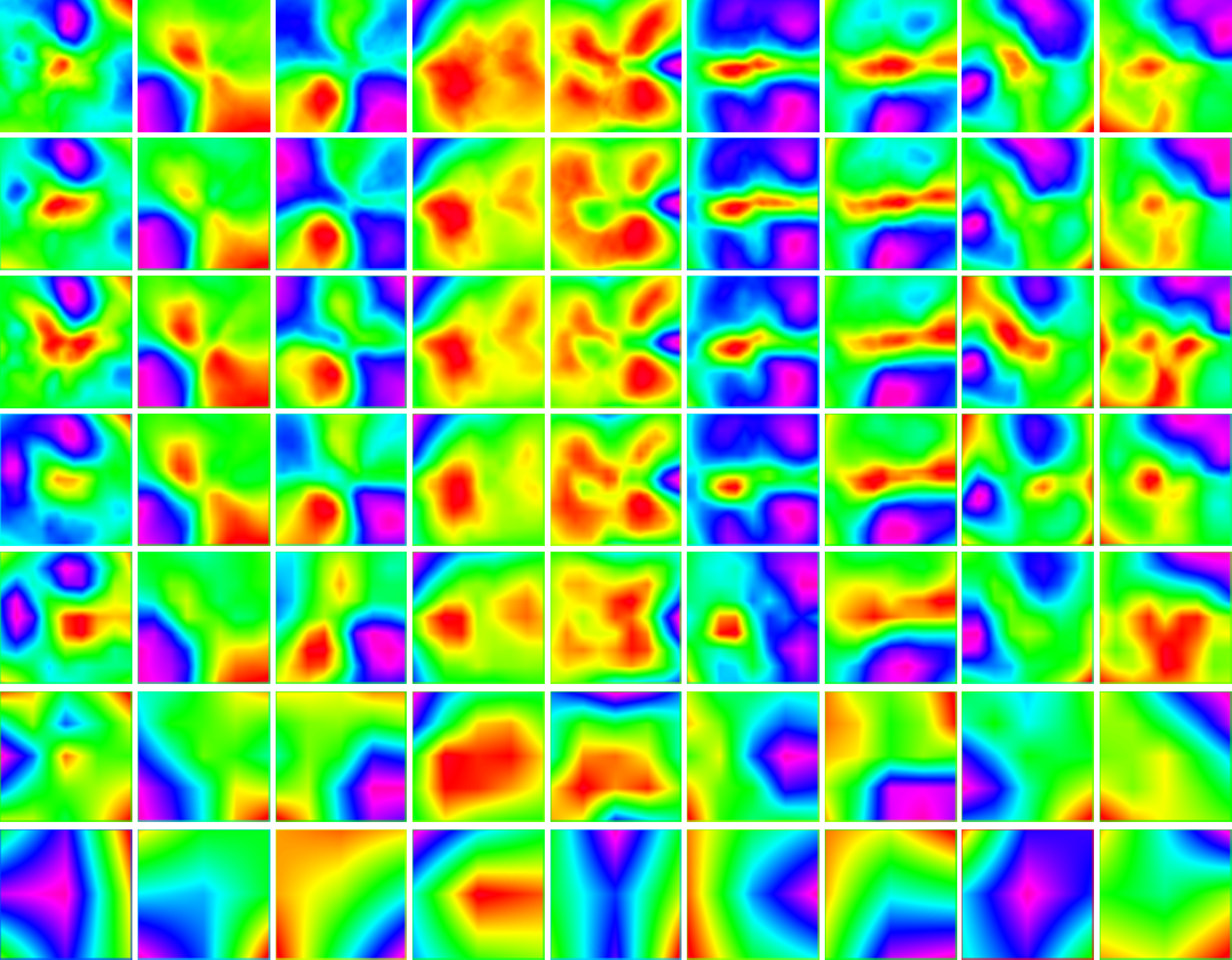}
\caption{
\textbf{LUTI$_{\text{uni}}$-MLP.} 
}
\label{fig:vis_emb_mlp_supp}
\end{subfigure}

\end{figure}
\clearpage   
\begin{figure}[tb]\ContinuedFloat
    
% second image
\begin{subfigure}[t]{1\columnwidth}
\centering
\includegraphics[width=1\columnwidth]{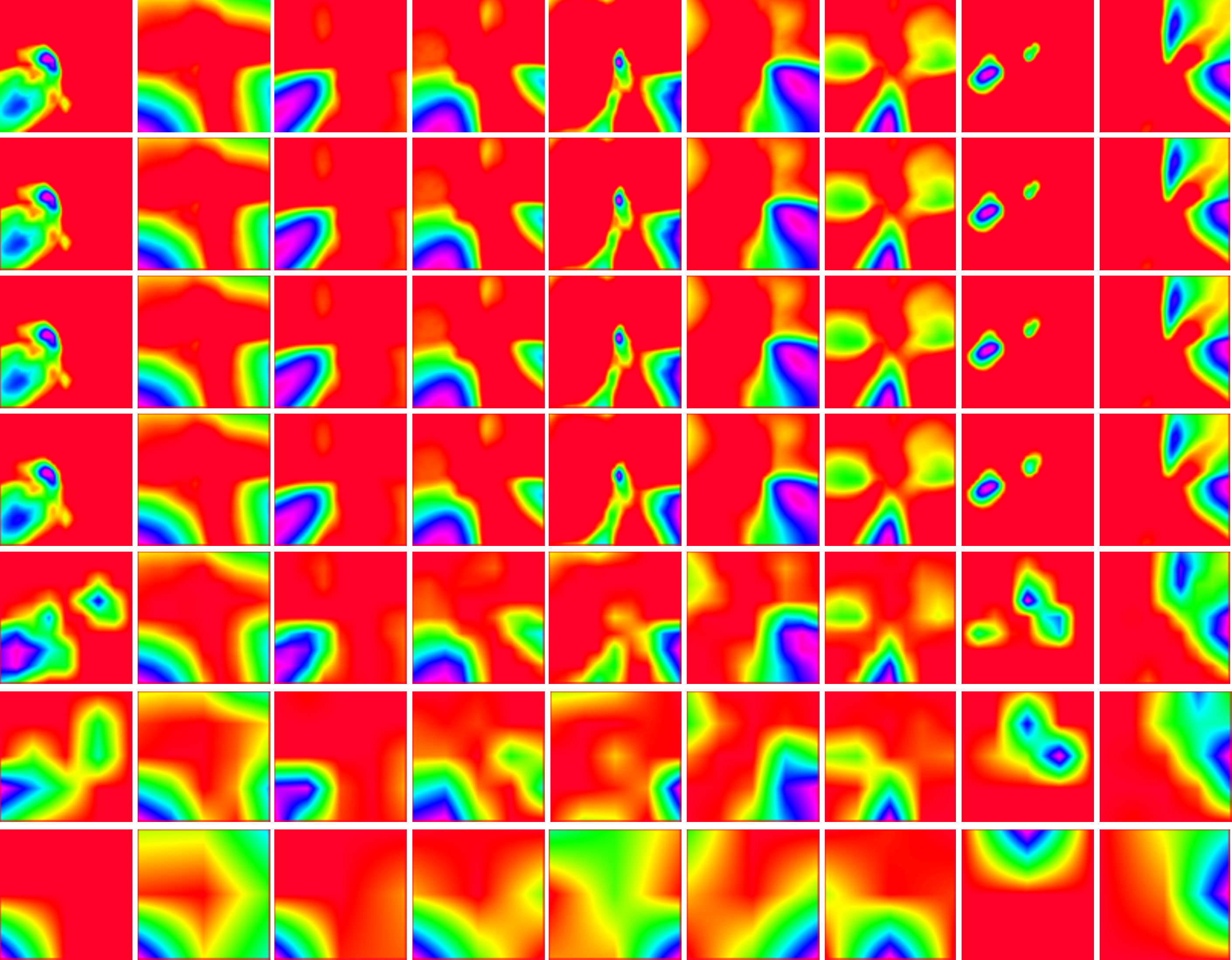}  
\caption{
\textbf{LUTI$_{\text{irr}}$-MLP.} 
}
\label{fig:vis_emb_mlp_irr_supp}
\end{subfigure}
\caption{
\label{fig:vis_emb_mlp_supp}
\textbf{Visualization of the trained embedding feature.}
From top to bottom, MLP (equivalent to PointNet), LUTI-MLP with discretizations $D=64,32, 16, 8, 4,$ and $2$.
Nine randomly selected channels of the slice on the $z=0$ plane are shown.
In LUTI$_{\text{uni}}$-MLP, the maximum is observed only on the edge. 
Conversely, we  can observe irregularly arranged peaks in case of $\text{LUTI}_{\text{irr}}$. 
Each LUTI variant used the same pre-trained model from MLP for $100$ epochs, using the ModelNet40 classification task
}
\label{fig:vs_emb}
\end{figure}
% Network is initialized from MLP trained 100 epoch and trained for another 100 epoch.

%%%%%%%%%%%%%%%%%%%%%%%%%%%%%%%%%%%%%%%%%%%%%%%
\newpage
\clearpage
\section{Visualization of the Embedding Space of LUTI-Direct}
\label{suppsec:vis_direct}
\Figure\ref{fig:vis_emb_direct_supp}, we show the trained embedding space that was directly trained for the table using LUTI$_{\text{uni}}$ without MLP (\Supp{\ref{suppsec:direct}}).
Compared with the variants trained with MLP (\Fig\ref{fig:vis_emb_mlp_supp}, top), the learned embedding feature of this variant tends to have more peaks when the lattice resolution is fine.
When $D$ is large, a large portion of the table is kept unchanged because the error signal rarely arrives (see \Supp{\ref{suppsec:direct}} and \Supp{\ref{suppsec:detail_analisys}} for the detailed analysis).
As $D$ decreases and the performance improves, the trained feature becomes smoother, and subjectively, it resembles the ones trained by MLP or LUTI-MLP shown in \Fig \ref{fig:vis_emb_mlp_supp}.

In this figure, we also present the results with different regularizations (\Supp{\ref{suppsec:direct}}).
Comparing the results with different regularizations, we can visually inspect and see that the one using TVL1 or TVL2 regularization tends to be smoother than the one without it.
\begin{figure}[!h]
\begin{subfigure}[t]{1.0\columnwidth}
\includegraphics[width=1\columnwidth]{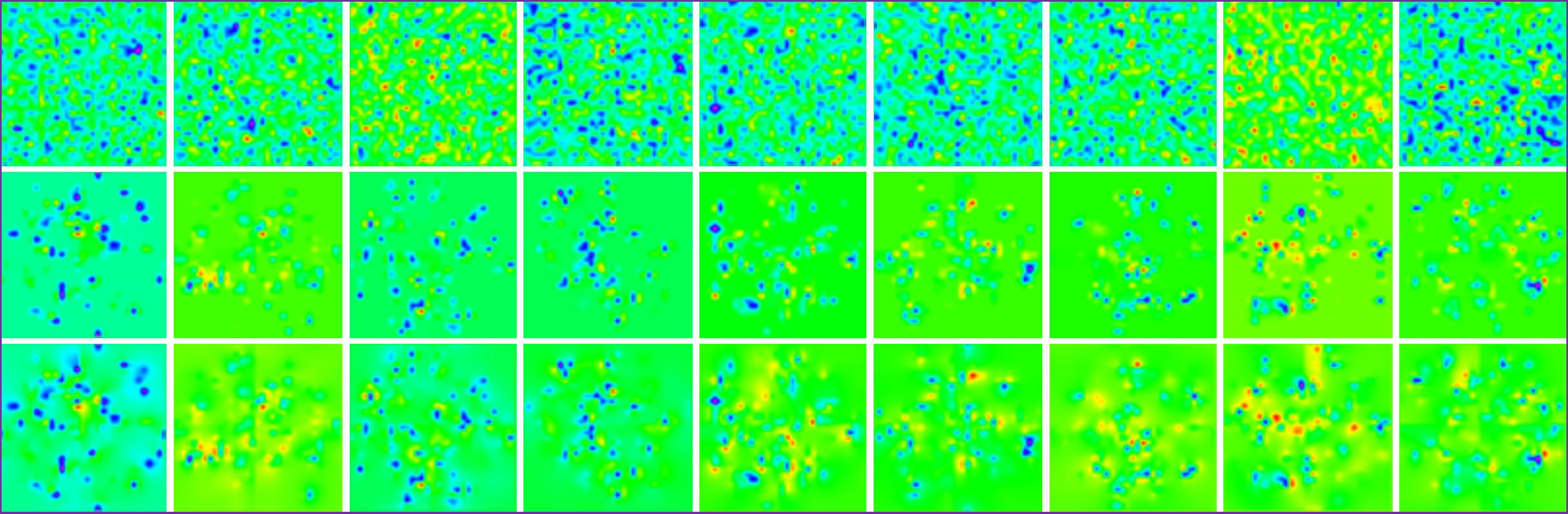}
\subcaption{$D=32$}
\end{subfigure}

\begin{subfigure}[t]{1.0\columnwidth}
\includegraphics[width=1\columnwidth]{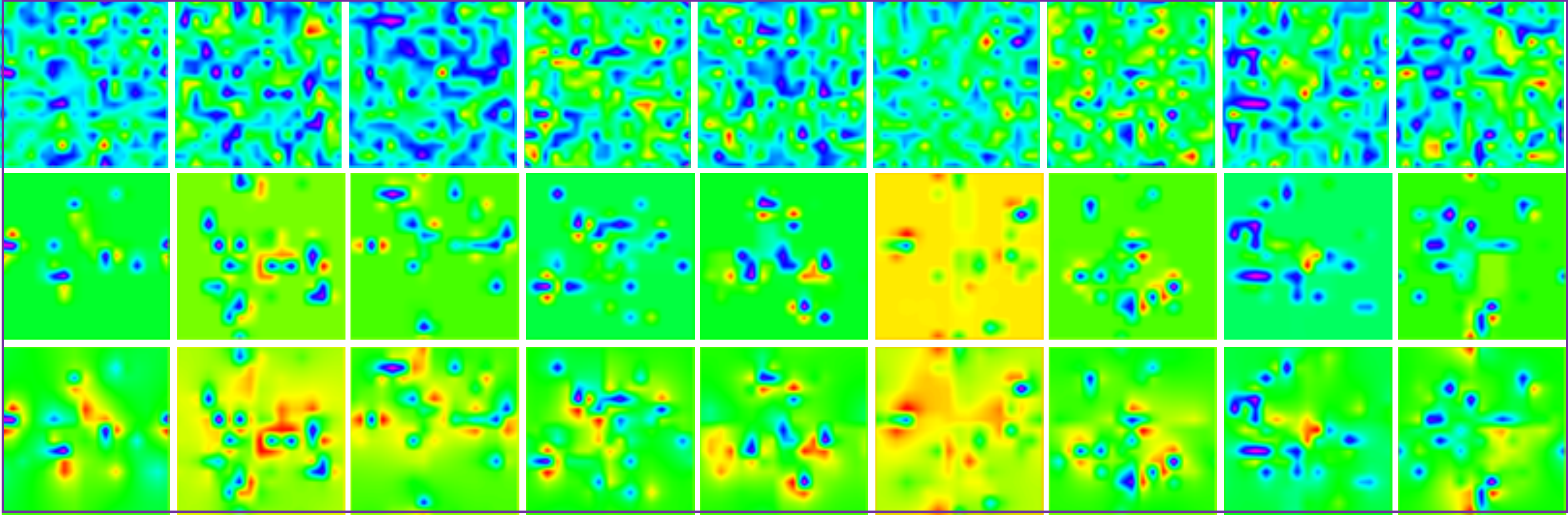}
\subcaption{$D=16$}
\end{subfigure}
\end{figure}
\clearpage

\begin{figure}[!h]\ContinuedFloat
\begin{subfigure}[t]{1.0\columnwidth}
\includegraphics[width=1\columnwidth]{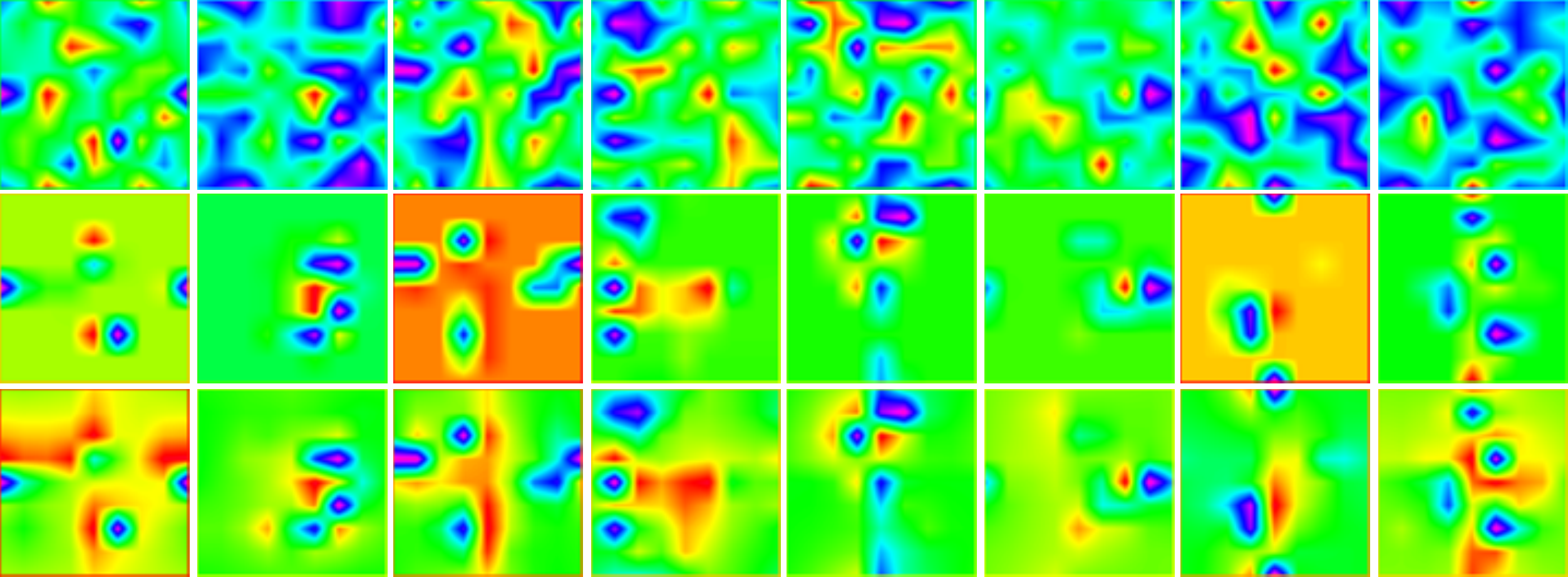}
\subcaption{$D=8$}
\end{subfigure}

\begin{subfigure}[t]{1.0\columnwidth}
\includegraphics[width=1\columnwidth]{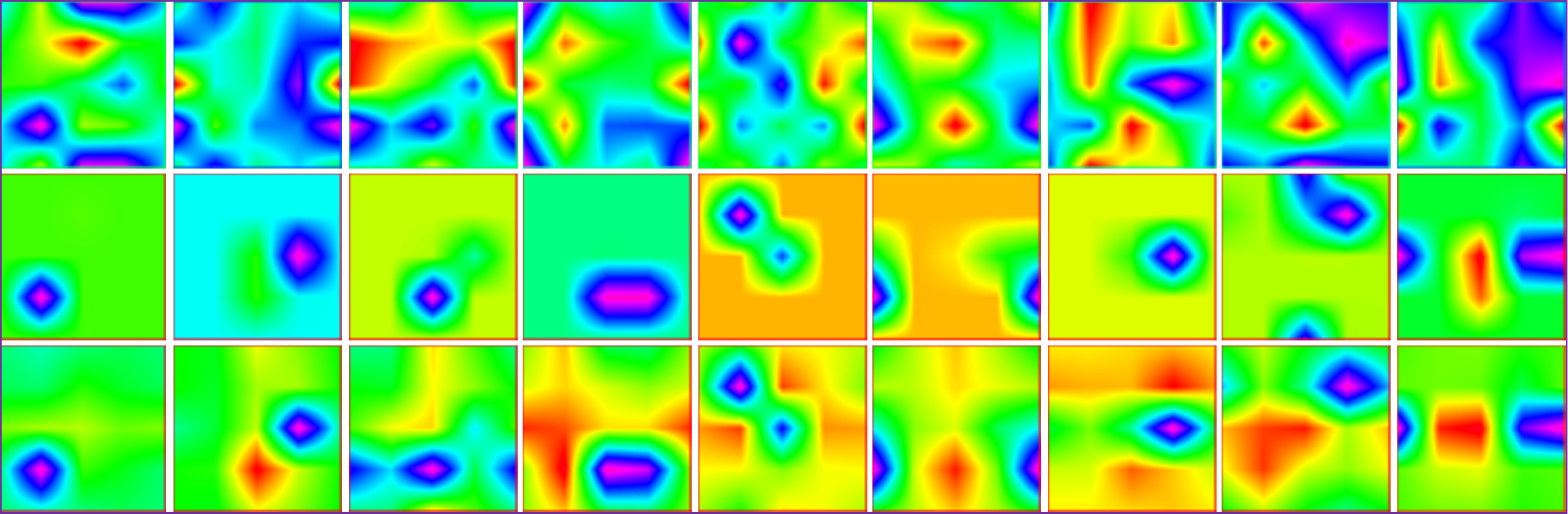}
\subcaption{$D=4$}
\end{subfigure}

\begin{subfigure}[t]{1.0\columnwidth}
\includegraphics[width=1\columnwidth]{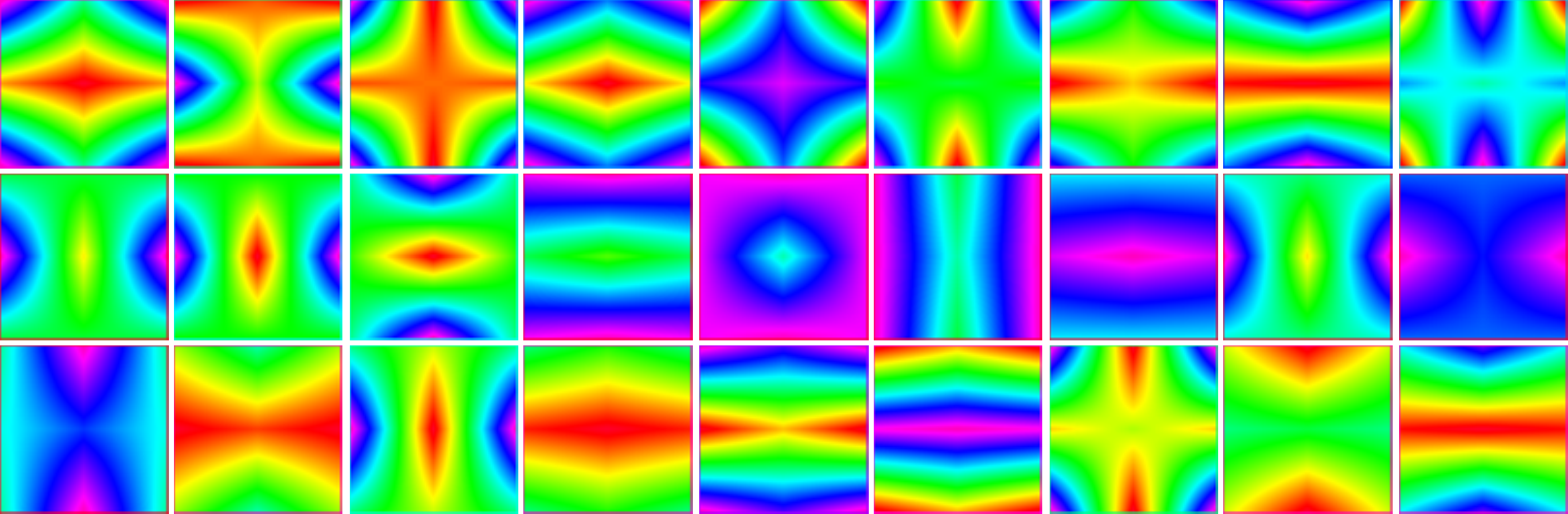}
\subcaption{$D=2$}
\end{subfigure}
\caption{
\label{fig:vis_emb_direct_supp}
\textbf{Visualization of the trained embedding feature of PointNet with direct table learning.}
% The trained embedding space from \textit{LUTI-Direct} is visualized.
The results are from different lattice resolutions $D$ and different regularizations.
The learned embedding space of LUTI-Direct with $D=32, 16, 8, 4,$ and $2$ are shown in sub-figure (a)-(e): for each  sub-figure, without regularization (top),  TVL1 regularization (middle), and TV L2 regularization (bottom).
Nine randomly selected channels of the slice on the $z=0$ plane are shown horizontally
}
\end{figure}

\end{document}